\documentclass[conference]{IEEEtran}
\ifCLASSOPTIONcompsoc
  \usepackage[nocompress]{cite}
\else
  \usepackage{cite}
\fi
\ifCLASSINFOpdf
\else
\fi

\usepackage{graphicx}
\usepackage{color,soul}
\graphicspath{{image/}}
\usepackage{setspace}
\usepackage{lmodern}
\usepackage{cite}
\usepackage[utf8x]{inputenc}
\usepackage[keeplastbox]{flushend}
\usepackage{floatrow}
\usepackage{amsmath,amsfonts,amsthm}
\usepackage{multirow}
\usepackage{mathptmx} 
\usepackage{amsmath}
\usepackage{algorithm}
\usepackage{algpseudocode}
\newcommand{\ignore}[1]{}
\usepackage{fancyhdr}
\usepackage[normalem]{ulem}
\usepackage[hyphens]{url}
\usepackage{microtype}
\usepackage{setspace}
\usepackage{balance}
\usepackage{subfig}
\usepackage{hyphenat}
\usepackage{array}
\usepackage{sidecap}
\usepackage[bookmarks=true,breaklinks=true,letterpaper=true,colorlinks,linkcolor=black,citecolor=blue,urlcolor=black]{hyperref}
\usepackage{hyperref}
  \hypersetup{%
    unicode=true,%
    pdfstartview={FitH},%
    colorlinks=true,%
    citecolor=blue,%
}

\algrenewcommand\algorithmicindent{1.0em}%
\algnewcommand\algorithmicto{\textbf{to}}
\algblockdefx[ForTo]{ForTo}{EndFor}
[3]{\algorithmicfor\ $#1 \gets #2$ \algorithmicto\ $#3$ \algorithmicdo}
{\algorithmicend\ \algorithmicfor}

\usepackage{tikz}
\usepackage{xcolor}
\newcommand*{\affaddr}[1]{#1} 
\newcommand*{\affmark}[1][*]{\textsuperscript{#1}}

\begin{document}
\bstctlcite{IEEEexample:BSTcontrol}
%
\title{Learning Forward Reuse Distance}
%
%
%
%

\author{Pengcheng Li\affmark[1] \hfil
        Yongbin Gu\affmark[2]
        \\
	 \affaddr{\affmark[1]Damo Academy of Alibaba Group Inc.}
     \affaddr{\affmark[2]Oregon State University}\\
    \IEEEauthorblockA{
	 pengcheng.li@alibaba-inc.com, guyo@oregonstate.edu}
	 }

\markboth{IEEE TRANSACTIONS ON COMPUTERS}%
{Shell \MakeLowercase{\textit{et al.}}: Bare Demo of IEEEtran.cls for Computer Society Journals}
%



\IEEEtitleabstractindextext{%
\begin{abstract}
Caching techniques are widely used in the era of cloud computing from applications, such as Web caches to infrastructures, Memcached and memory caches in computer architectures. Prediction of cached data can greatly help improve cache management and performance. The recent advancement of deep learning techniques enables the design of novel intelligent cache replacement policies.

In this work, we propose a learning-aided approach to predict future data accesses. We find that a powerful LSTM-based recurrent neural network model can provide high prediction accuracy based on only a cache trace as input. The high accuracy results from a carefully crafted locality-driven feature design. Inspired by the high prediction accuracy, we propose a pseudo OPT policy and evaluate it upon 13 real-world storage workloads from Microsoft Research. Results demonstrate that the new cache policy improves the state-of-art practical policies by up to 19.2\% and incurs only 2.3\% higher miss ratio than OPT on average.
\end{abstract}

}

\maketitle

\IEEEdisplaynontitleabstractindextext

%


\section{Introduction}
\label{sec:intro}
The exponential increase in computing capabilities in hardware has greatly advanced various applications of deep neural networks, such as speech recognition~\cite{Hinton+:2012,GravesH:arxiv13}, visual object recognition~\cite{Krizhevsky+:NIPS17,Howard+:arxiv17}, and natural language processing~\cite{Cho+:arxiv14,Sutskever+:NIPS14}. 
The great success of these real-world applications has inspired this line of research using deep neural networks upon classical research problems.

 
Cache is a universal and indispensable component in computer science. In a computer, a CPU cache addresses the memory wall~\cite{WulfM:1995} problem in von Neumann computers: Computation is orders of magnitude faster than memory access. In a Web server, accessing back stores suffers a long latency due to low network bandwidth. A Web server mitigates this problem by caching frequently accessed items in a Web cache. 1\% increase of cache hit rate could result in 35\% reduction of latency~\cite{Cidon+:NSDI16}. In content delivery networks, cache storage at routers becomes of utmost importance to handle the exponential increase in video streaming services~\cite{Narayanan+:NETAI18,Berger+:NSDI17}. To implement a cache with ultra-high performance, the data locality characterization of programs becomes critical. The target cache in this paper is a general cache in theory.

\begin{figure}[t]
\centering
\includegraphics[width=0.8\linewidth]{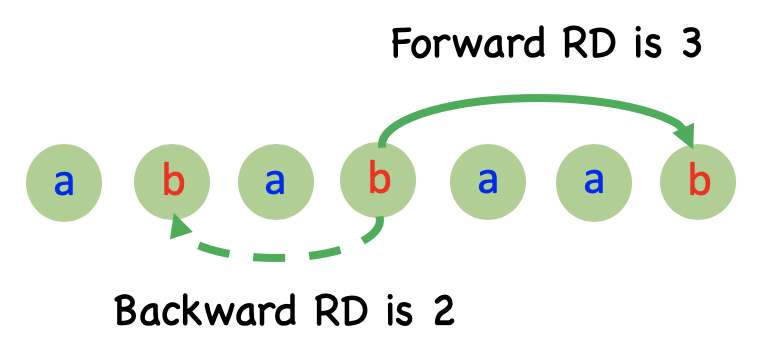}
\caption{An example of reuse distance. The solid arrow shows forward reuse distance. The dash arrow shows backward reuse distance.}
\label{fig:reuse-distance}\vspace{-10pt}
\end{figure}

Peter Denning, who found the working-set theory~\cite{Denning:CACM68}, defined the \textit{principle of locality} as ``the tendency for programs to cluster references to subsets of address space for extended periods,'' called phases~\cite{DenningM:Book15}. \textit{Reuse distance} is the de facto metric to quantify data locality in a program. In this paper, we define reuse distance as ``the number of accesses between two consecutive accesses to the same datum''~\footnote{Other works define it as ``the number of unique accesses between two consecutive accesses to the same datum''. For example, stack distance is called reuse distance.}. The definition is trace-based. A cache trace is a sequence of cache accesses. In a CPU cache, a trace is a memory access trace; in a Web cache, a trace is a series of Internet data objects. 
There are two types of reuse distance, backward reuse distance and forward reuse distance, as shown in Figure~\ref{fig:reuse-distance}. The second access of ``b'' has forward reuse distance of 3 and backward reuse distance of 2. The backward reuse distance is often referred to as reuse distance for simplicity. The first access of any data has $\infty$ reuse distance; the last access has $\infty$ forward reuse distance. 

Being aware of forward reuse distance on-line is greatly beneficial. The Belady's optimal algorithm~\cite{Belady66} (OPT) discards the data that will not be needed for the longest time in the future for a cache eviction event. Forward reuse distance just foresees the future access time. By it, we can derive the eviction candidate at an eviction event for implementing the OPT policy. To be aware of forward reuse distance for all data accesses is theoretically equivalent to know of the future.
The online computation is easy for reuse distance resulting from the past accesses (explained in Section~\ref{sec:back-reuse-distance}), but is impossible for the forward reuse distance, because the future has not been ``seen'' yet. 

This paper strives to predict forward reuse distance based on only the past cache trace by applying a novel learning-aided approach. In addition, we demonstrate the strength of our approach by leveraging it to advance the state-of-the-art in the area of cache replacement.


Recently, the compute architecture community has applied machine learning techniques to predict the next memory address for the CPU caches~\cite{Hashemi+:ICML18,Shi+:MICRO19} with the assistance of \textit{program context such as program counters}. In particular, program counters are a dominant factor in prior work~\cite{Shi+:MICRO19}, and they format a sequence of pairs of program counter and memory address as a trace.  As a result, their methods cannot adapt to other use cases that do not have program counters (such as Web caches). 

\textbf{Novelty.} In this paper, we study the prediction based on \textit{only a cache trace}, which is built on a theoretical foundation that a cache trace is not a random sequence of accesses, and instead contains locality insights inside~\cite{DenningM:Book15}. Any cache replacement policy would otherwise be useless. We will discuss cache trace patterns in Section~\ref{sec:cache-ptn}. \textit{No any prior arts ever made predictions based on only a cache trace.} To tackle the challenge, we take advantage of the powerful intelligent capability of deep neural networks to understand cache traces, by integrating locality insights in the model design.

In summary, this paper makes the following contributions:
\begin{itemize}
\item We design a powerful, unconstrained multi-LSTM recurrent neural network (RNN), inspired by recent success of long-short-term-memory (LSTM) RNN\cite{HochreiterS:1997,HochreiterS:NIPS96} in sequence modeling tasks~\cite{Sutskever+:NIPS14,Shi+:MICRO19,Cho+:arxiv14}, as the first work to apply deep learning to predict forward reuse distance and demonstrate its high accuracy and availability.
\item We explore to make use of a set of locality intrinsic features in the training design to verify that the prediction based on only a cache trace is highly viable. Therefore, the proposed techniques are widely applicable for extensive cache application scenarios.
\item We propose to utilize the classical \textit{K-means} algorithm to cluster data address deltas in the training process to improve prediction accuracy.
\item We present a novel prediction-based pseudo OPT cache replacement policy, inspired by the good prediction results by the proposed neural network model. 
\item We measure our model and the new cache replacement policy on 13 Microsoft storage traces, which reveal a broad range of cache access patterns. Results demonstrate that our method improves the state-of-art practical policies by up to 19.2\% and incurs only 2.3\% higher miss ratio than OPT on average.
\end{itemize}

\section{Background}
This section describes the concepts of reuse distance and recurrent neural networks.

\subsection{Reuse Distance}
\label{sec:back-reuse-distance}
Mattson et al.~\cite{Mattson+:IBM70} defined the well-known stack distance as the number of \textit{unique} accesses between two consecutive accesses to the same data block. Stack distance is yet different than reuse distance referred to by this paper. For example, given a
sequence of memory accesses: $a_1$, $b_1$, $c_1$, $b_2$, $a_2$, where the subscripts represent the access number of the same data block. The stack distance for data block $a$ at the access instance $a_2$ is $2$ since two other data blocks $b$ and $c$ are accessed between consecutive access to $a$. The reuse distance of $a_2$ is 4, since there are 4 accesses in between.
Reuse distance and stack distance are convertible to each other. 

Stack distance can be used to directly compute the cache miss ratios for all cache sizes~\cite{Zhong+:PLDI04}, which is the well-known miss ratio curve (MRC). Like stack distance, reuse distance can also be used to compute the MRC via a new locality metric, namely \textit{volume fill time} by Xiang et al.~\cite{Xiang+:ASPLOS13}. In general, the volume fill time is the average time for a program to fill the cache of some size. If reuse distance is larger than the volume fill time, then the corresponding access is a miss. By counting total misses through the volume fill time metric, the miss ratio can be easily computed.

The volume fill time is easier to calculate as the computation of reuse distance can be done within linear time, while the previous stack distance takes $nlog(n)$ time complexity~\cite{Zhong+:PLDI04}, where $n$ is the length of the cache trace. The reuse distance for all accesses can be computed in one pass: We may use a hash map to record the last access time for each data block, the reused distance is computed as the current time minus its last access time.

\subsection{Recurrent Neural Networks}
Recurrent neural networks have achieved great success on solving sequential prediction problems, such as speech recognition~\cite{Hinton+:2012,GravesH:arxiv13}, visual object recognition~\cite{Krizhevsky+:NIPS17,Howard+:arxiv17}, and natural language processing~\cite{Cho+:arxiv14,Sutskever+:NIPS14}. Both Long Short Term Memory (LSTM)~\cite{Chung+:arxiv14} and Gated Recurrent Unit (GRU)~\cite{Chung+:arxiv14} are popular models of recurrent neural networks. In this paper, we employ the LSTM model to predict reuse distance as it is good at dealing with long sequences by propagating the internal hidden state additively instead of multiplicatively.


An LSTM unit is composed of a hidden state $h$, a cell state $c$, an input gate $i$, a forget gate $f$, and an output gate $o$, which enable it to obtain and store the information from the input, and propagate it to the next time point. At time $t$, the LSTM is input with $x_t$, and the LSTM states are computed using the following formula:

\begin{enumerate}
\item Compute the input, forget, and output gates
  \begin{equation}
  \label{eq:lstm-eq1}
    \begin{aligned}
    i_t &= \sigma(W_i\text{[}x_t, h_{t-1}\text{]} + b_i) \\
    f_t &= \sigma(W_f\text{[}x_t, h_{t-1}\text{]} + b_f) \\
    o_t &= \sigma(W_o\text{[}x_t, h_{t-1}\text{]} + b_o)
    \end{aligned}
  \end{equation}
\item Update the cell state
  \begin{equation}
  \label{eq:lstm-eq2}
    \begin{aligned}
    c_t &= f_t \odot c_{t-1} + i_t \odot \tanh(W_c\text{[}x_t, h_{t-1}\text{]} + b_c)
    \end{aligned}
  \end{equation}
\item Compute the LSTM hidden (output) state
  \begin{equation}
  \label{eq:lstm-eq3}
    \begin{aligned}
    h_t = o_t \odot \tanh(c_t)
    \end{aligned}
  \end{equation}
\end{enumerate}

\noindent
Where [x$_t$, h$_{t-1}$] denotes the concatenation of the current input and previous hidden state, $\odot$ denotes element-wise multiplication, and $\sigma(u) = \frac{1}{1+\exp{-u}}$ is the sigmoid method.

The above process depicts the working mechanisms of a single LSTM cell. A complete LSTM model can be comprised of tens of the cells by cascading them together. The cell numbers are called \textit{LSTM width}. LSTM is good at memorizing long-term sequences, and with the deployment of multiple layers it can also be used to map from sequence to sequence~\cite{Sutskever+:NIPS14}. 
Caching problems are much like sequence problems that have been solved.
Therefore, we employed a LSTM-based RNN that has multiple layers to learn forward reuse distance in this paper.

\section{Cache Trace Patterns}
\label{sec:cache-ptn}
Denning defined \textit{locality} as ``the tendency for programs to cluster references to subsets of address space for extended periods''~\cite{Denning:CACM68,DenningM:Book15}. Locality has the implications of patterns and phases. We have analyzed 13 storage traces from Microsoft Research~\cite{Narayanan+:TOS08}. These traces represent different workload behaviors happening in the Microsoft Cloud. 


\begin{figure*}[t]
\centering
\subfloat[Triangle]{
\includegraphics[width=0.33\textwidth]{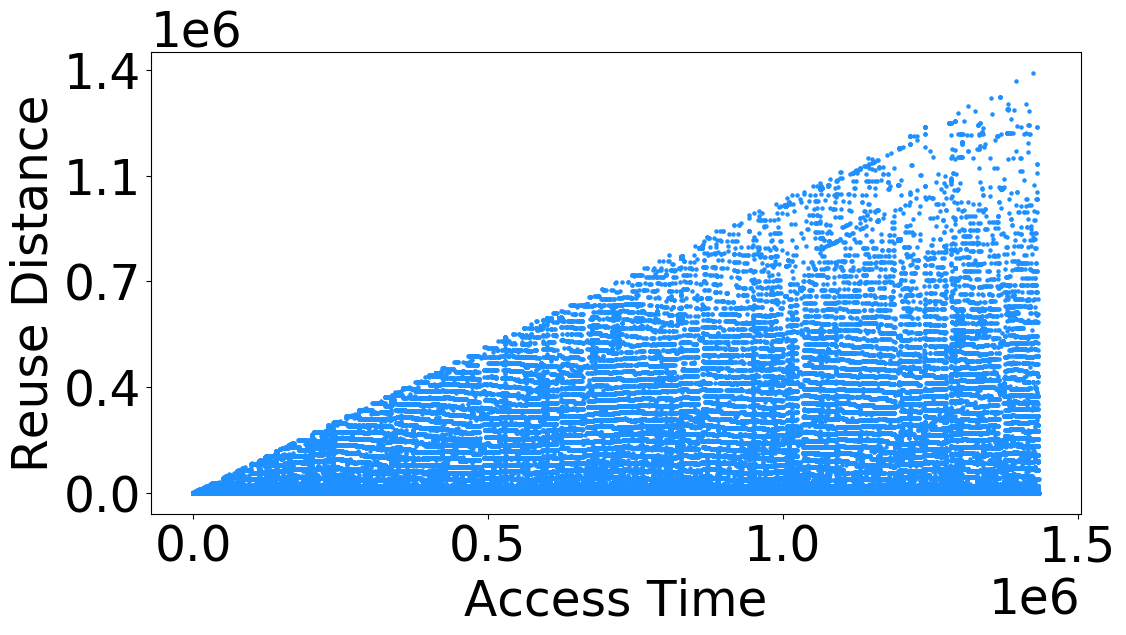}
}
\subfloat[Clouds]{
\includegraphics[width=0.33\textwidth]{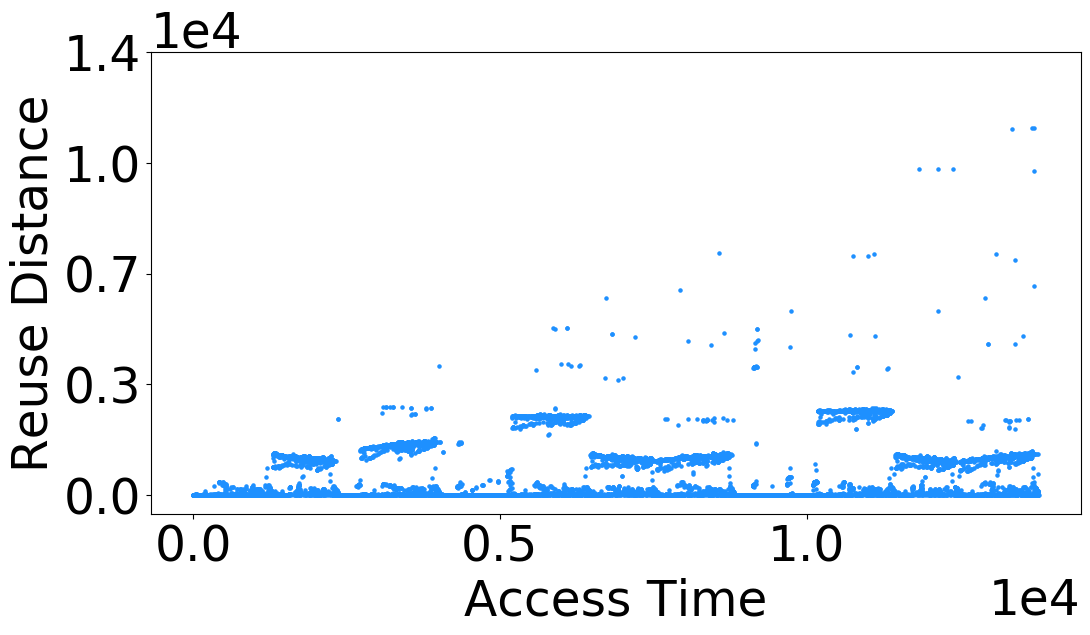}
}
\subfloat[Gaps and bars]{
\includegraphics[width=0.33\textwidth]{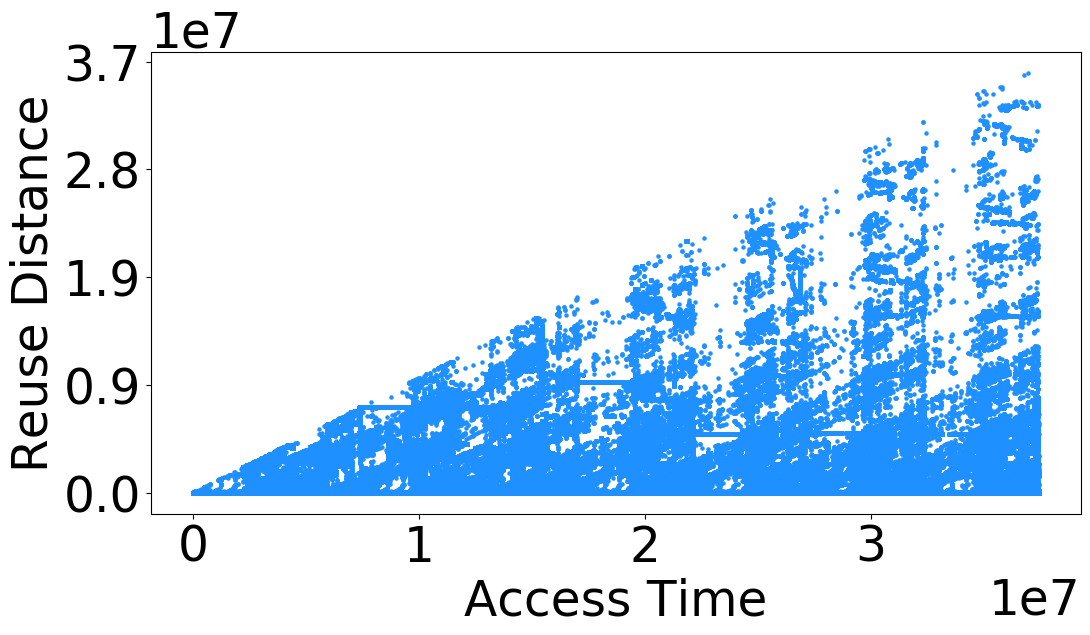}
}
\\
\subfloat[Crossing lines]{
\includegraphics[width=0.33\textwidth]{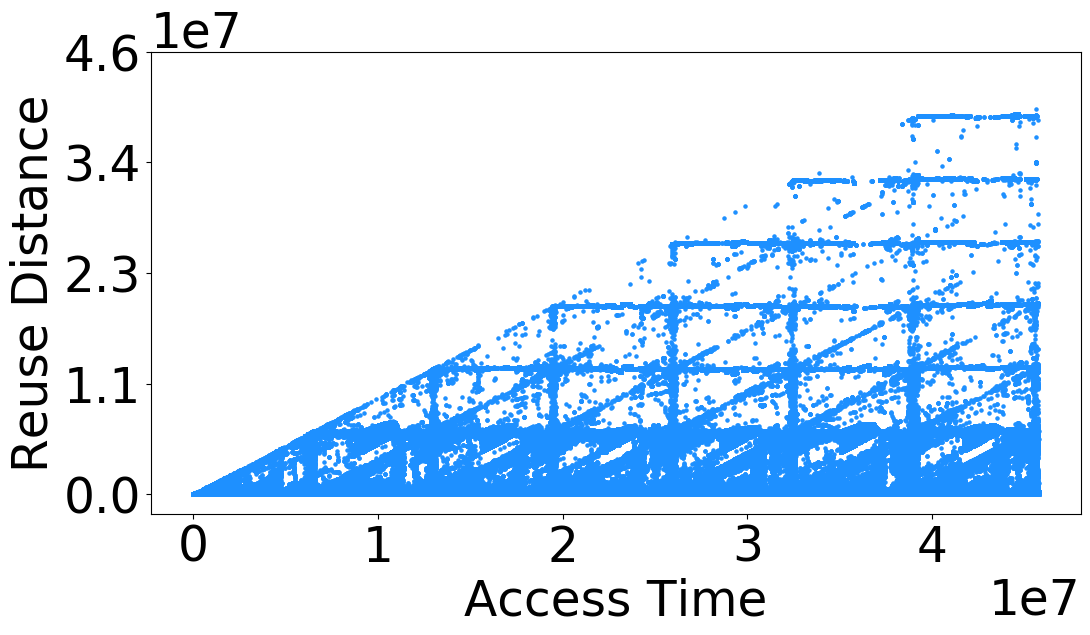}
}
\subfloat[Horizontal lines]{
\includegraphics[width=0.33\textwidth]{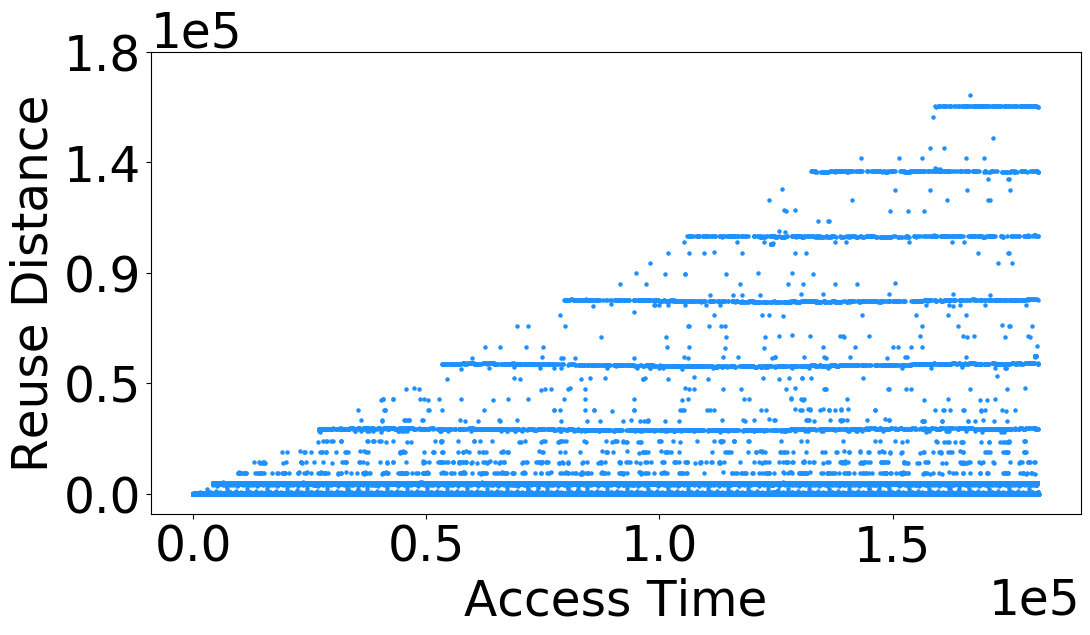}
}
\subfloat[Clusters]{
\includegraphics[width=0.33\textwidth]{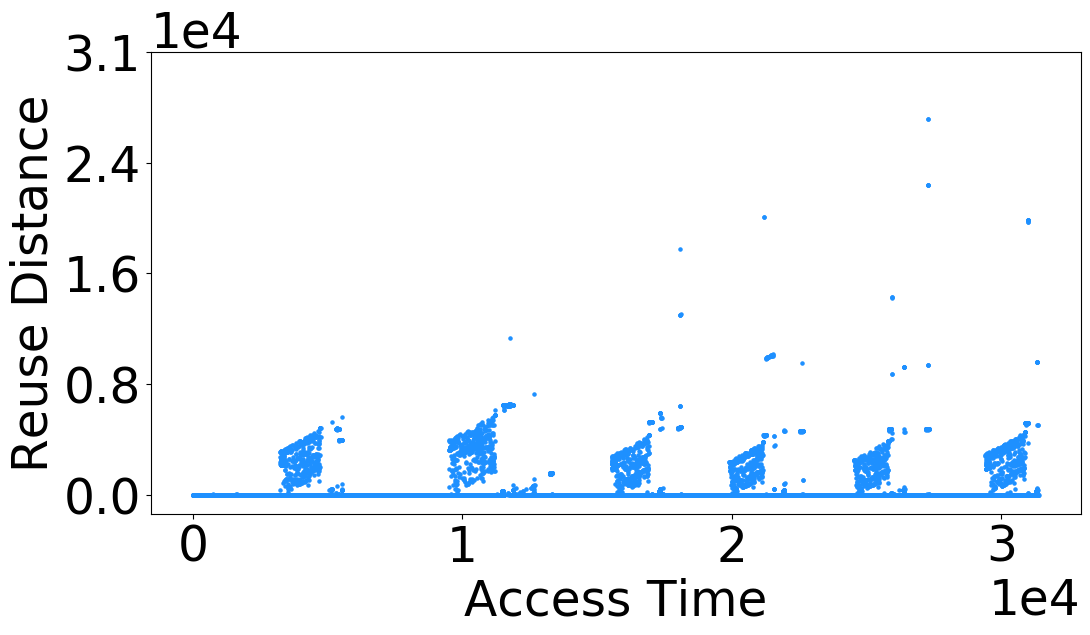}
}
\caption{13 MSR storage traces demonstrate different patterns. We filter out those traces that do not have an observable pattern and classify the remaining into six types of pattern. For the purpose of presentation, we use $0$ to denote $\infty$ reuse distance.}
\label{fig:trace-pattern}
\end{figure*}

We characterize the patterns of these traces on a basis of time-series reuse distance trend, and classify these traces into six types of pattern, which are shown in Figure~\ref{fig:trace-pattern}. In the figure, x-axis uses logical time, which starts from $0$ and is incremented by $1$ at each access. 
The y-axis is the reuse distance of the data access.
The first access for any data has $\infty$ reuse distance. For ease of presentation, we present $\infty$ reuse distance with $0$ in the figure. The bottom line thus (where y-axis equals to 0) shows the number of unique data.


The first pattern is a ``triangle''. The slope line shows that the maximum of reuse distance is getting larger as more data is accessed over time. Under this pattern, any single datum could possibly be reused. Along the slope line bottom-up, the point density is getting smaller, indicating that most of the data have relatively short reuse  distance.


We define the second pattern as a ``clouds'' pattern. From a visual point of view, reuse distances form many clouds. In each piece of cloud, continuous accesses reuse the data from early accesses with similar distance. 
This pattern is a good example to demonstrate the definition of locality. 

We refer to the third pattern as ``gaps and bars''. In ``gap'' areas, data accesses have short reuse distance; in ``bars'' areas, a single data access may reuse the data from any earlier access so that the reuse distance could be either long or short. This pattern exhibits an interleaving phase behavior. Same as the first pattern, it has a slope line showing that the maximum of reuse distance is getting larger over time.

The fourth pattern is ``crossing lines''. There are three kinds of line: horizontal, vertical and slope. The horizontal lines show that there are constantly data accesses having the same reuse distance. Vertical lines show that close accesses have a broad range of reuse distance. The slope lines show that continuous accesses reuse with steadily increasing distance. 

The fifth pattern is mainly composed of horizontal lines. Most reuse distances fall within a small number of values. Similar as the fourth pattern, these lines show that continuous accesses reuse earlier accesses with the same distance.

We refer to the sixth pattern as ``clusters''. The traces with this pattern periodically access data with a "clustering" reuse distance. Except these clusters, other data accesses have very short reuse distance, or are even not reused. 


In this paper, we explore to predict forward reuse distance for cache traces that exhibit patterns or phases, i.e., locality by employing an LSTM-based RNN. 


\section{RNN Model Design}
\label{sec:lstm}
RNNs have shown their great success in the sequence-to-sequence machine translation problem \cite{Cho+:arxiv14}, and the prediction of forward reuse distance can be considered as a sequence-to-sequence problem~\cite{Sutskever+:NIPS14}. Consequently, we use the long short term memory (LSTM) networks to prediction forward reuse distance. 


\subsection{Multi-LSTM RNN}

\begin{figure}[t]
\centering
\includegraphics[width=0.95\textwidth]{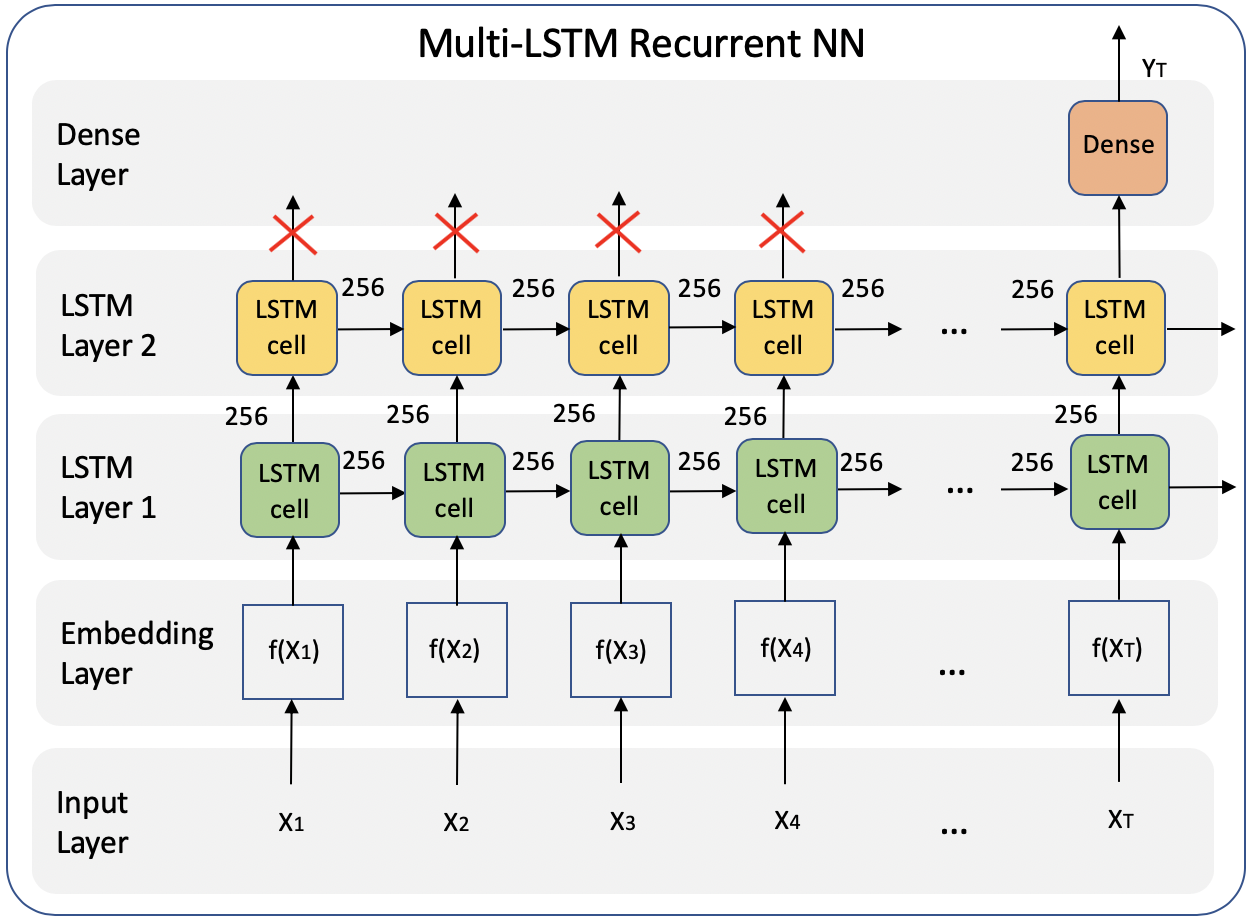}
\caption{The Multi-LSTM RNN used in our work. $x_i$, where $i \in 1 \dots T$, denotes a feature tensor. $x_i$ is embedded. Given a sequence of input features \{$x_1$, $x_2$, $\dots$, $x_T$\}, we have done predicting for time $T-1$ and are predicting for time $T$. We feed \{$x_1$, $x_2$, $\dots$, $x_T$\} into the network. The network has two LSTM layers. After two-layer network training, it outputs a 256-dimensional tensor. Finally, the network uses a dense layer to output a single floating-point value. LSTM width is 256 for an example. The outputs for \{$x_1$, $x_2$, $\dots$, $x_{T-1}$\} are not used.}
\label{fig:encoder-decoder}
\end{figure}

We use a stacked LSTM-based RNN~\cite{Sutskever+:NIPS14,Cho+:arxiv14} model, which is composed of multiple LSTM layers. 
An LSTM layer reads in a sequence of feature tensors entirely, called a \textit{sample}. We call its length as \textit{sequence\_length} in this paper. 
Generally, an LSTM layer processes a sample of length of \textit{sequence\_length} step by step. In each step, it processes an one-dimensional feature tensor by reading in a previous hidden state as well as an optional output of previous step, and then generates an output tensor and a hidden state for the next step to use. The dimension of the output tensor is the same as the width of an LSTM cell.
Once the entire sequence of feature tensors is processed, a layer generates a new sequence of tensors by composing the output tensors of all steps into a sequence, and feeds the sequence to the next layer. The next layer repeats the above process unless it is the last layer. The last layer executes the same way as previous layers except that it only outputs a tensor in the last step. Finally, the model applies a dense layer to compress the tensor to a single value as the prediction. 


Figure~\ref{fig:encoder-decoder} shows the proposed model. In the network, we use LSTM to capture long-term data dependency among the sequential input data. LSTM has a special design property related to carefully avoiding vanishing and exploding gradient problem when building deep layer neural network models~\cite{Zekany+:MICRO16}. Gated Recurrent Unit (GRU) is another model that captures long-term data dependency~\cite{Chung+:arxiv14}. Chung et al.~\cite{Chung+:arxiv14} showed that LSTM and GRU had comparable performance.

In Figure~\ref{fig:encoder-decoder}, we are predicting forward reuse distance for time $T$. The input is a sequence of feature tensors \{$X_1$, $X_2$, $\dots$, $X_T$\}, where $X_i$, $i \in 1 \dots T$, denotes a feature tensor, which is of floating-point values. We first scale all feature values down to [-1, 1] through an embedding layer and then feed a sequence of these tensors to two LSTM layers. LSTM has a certain number of cells inside. The number is called \textit{LSTM\_width}. 
Intermediate temporary tensors, such as hidden states or temporary output tensors, are all of width of \textit{LSTM\_width}. Figure~\ref{fig:encoder-decoder} uses 256 for example. Finally, the model
generates an output, which is a 256-dimensional tensor, and then uses a dense layer to convert it to one floating-point value $Y_{T}$.


For high prediction accuracy, in addition to features, we also make the target value, i.e., predicted forward reuse distance, scaled down to [-1, 1]~\footnote{Assume $M$ denotes the maximal target forward reuse distance and $N$ the minimum. Given any forward reuse distance $RD_i$, the scaled value is $-1 + \frac{2}{M-N} \times (RD_i - N)$.}. Therefore, $Y_{T}$ is in [-1, 1]. We scale $Y_{T}$ back for real use.

For good prediction performance, in addition to a well-designed neural network, another key is the design of features. Next, we will introduce our locality-driven feature design.


\subsection{Locality-Driven Feature Design}
\label{sec:fea-design}

We extract features from a given cache trace. Each access in a cache trace corresponds to a feature tensor. Existing cache replacement policies, such as \texttt{LRU}, \texttt{LFU} or \texttt{ARC}~\cite{MegiddoM:FAST03}, consider either data recency or data frequency or both in design. 
\texttt{LRU} captures {data recency} by caching the most recent frequently accessed data blocks. \texttt{LFU} makes use of no data recency but data frequency, thus it may accumulate data blocks in cache that are frequently accessed.

We design the following features:

\vspace{0.75\baselineskip}
\noindent {\bf \textit{Data block address.}}
The first feature is data address or data block ID. For a CPU cache, it is a memory address. A problem is that the address range could be extremely broad. For example, in 64-bit machines, the memory address range is extremely large. Existing works~\cite{Hashemi+:ICML18,Shi+:MICRO19} have shown that LSTM has a poor capability in training when the number of unique address is very large. Therefore, we choose to use address delta~\cite{Hashemi+:ICML18} to represent address information. For each access, we compute the difference of the current address and the previous one. The first address is considered to be zero.
The unique address-deltas becomes fewer. Table~\ref{tbl:addr-delta} shows their compression ratios. For all tests except \textit{rsrch} and \textit{stg}, the unique numbers are tremendously reduced. In real, it is possible that the number of unique addresses is actually smaller but that of unique address-deltas is large, just as \textit{rsrch} and \textit{stg} show. We consistently use address deltas in our tests.

\begin{table}
\centering
\caption{Compression ratios when using address deltas. Negatives denote the compression gain. }
\begin{tabular}{|c|c|c|c|c|c|}
\hline
\textbf{Test} & \textbf{Ratio} & \textbf{Test} & \textbf{Ratio} & \textbf{Test} & \textbf{Ratio} \\\hline\hline
\textit{hm} & -96.20\%  & \textit{proj} & -92.50\% &
\textit{prxy} & -99.35\% \\\hline
\textit{mds} & -84.59\% & \textit{rsrch} & +526\% &
\textit{src1} & -13.50\%  \\\hline 
\textit{prn} & -98.94\% & \textit{src2} & -68.97\% &
\textit{stg} & +94\% \\\hline 
\textit{ts} & -68.05\% & \textit{usr} & -58.17\% &
\textit{wdev} & -73.07\% \\\hline 
\end{tabular}
\label{tbl:addr-delta}
\end{table}

\begin{figure}[t]
\centering
\scalebox{0.9}{
\includegraphics[width=1.1\linewidth]{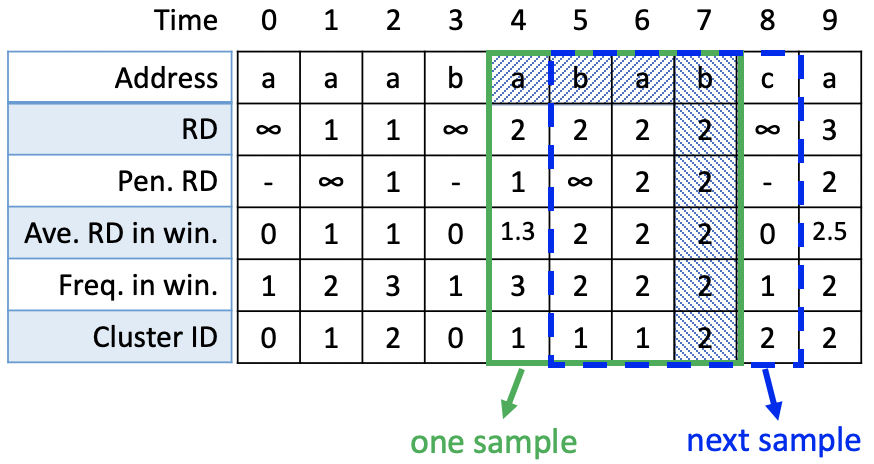}
}
\caption{How a data set is generated. The input cache trace is $aaabababca$. The feature tensor at time $7$ is $(b, 2, 2, 2, 2, 2)$. Assume that \textit{sequence\_length} equals to $4$. The green box frames the sample at time $7$, $((a, 2, 1, 1.3, 3,1)$, $(b, 2, \infty, 2, 2, 1)$, $(a, 2, 2, 2, 2, 1)$, $(b, 2, 2, 2, 2, 2))$. ``Ave. RD in win.'' denotes the average reuse distance in the sliding window of length 4. ``Freq. in win.'' denotes the frequency.}
\label{fig:lstm-example}
\end{figure}

\vspace{0.75\baselineskip}
\noindent {\bf \textit{Reuse distance.}} We use reuse distance as the second feature. Reuse distance captures data recency information. The computation of reuse distance is on-line by using a hash table to record the last access time for each data. Section~\ref{sec:back-reuse-distance} already discussed its algorithm. Taken as example, Figure~\ref{fig:lstm-example} shows a cache trace $aaabababca$. The second row denotes the reuse-distance feature.

\vspace{0.75\baselineskip}
\noindent {\bf \textit{Penultimate reuse distance.}} 
The penultimate reuse distance is referred to as the third feature. It is also a data recency feature. This feature strengths data recency information in addition to reuse distance. For computation, we just need to copy reuse distance of last access to the current access. It can be illustrated in Figure~\ref{fig:lstm-example} by referring to the third row in the table.

\vspace{0.75\baselineskip}
\noindent {\bf \textit{Average reuse distance in the sliding window.}} 
Reuse distance and penultimate reuse distance are both individual accounting for a data block. This feature offers expected reuse distance for a data block in the past $k$ accesses. For a data block, we count finite reuse distance and compute the average.
The parameter $k$ controls how far we would look back to history. 
For computation, we use a sliding window algorithm to compute it on-line in $O(1)$ time complexity. The sliding window moves to next access by dropping reuse distance on the left end of the window and count in that on the right end. We can use a cyclic buffer to record reuse distance at each time point.

A comparable feature is the average reuse distance over the entire history. This is an extreme case in our feature when $k=\infty$. Although this gives an expected estimation of next possible reuse distance, it loses data recency information. We claim that it is useful to approximate the average forward reuse distance. However, we need to predict for a specific forward reuse distance. Hence, the parameter $k$ is preferred to be small. We choose $100$ in this paper.

\vspace{0.75\baselineskip}
\noindent {\bf \textit{Frequency in the sliding window.}} 
For the current data block, we count its access frequency in the past $k$ accesses. This feature is a data frequency feature. In this paper, we set $k$ with 50, empirically.


\vspace{0.75\baselineskip}
\noindent {\bf \textit{Example.}}  Figure~\ref{fig:lstm-example}
gives an example to show how we compute these features. The input is a trace in the first row. We use letters to denote a address. Look at the access to ``b'' at time 5. Its reuse distance is 2 and penultimate reuse distance is $\infty$. Looking back to the last 4 accesses from the ``b'', there is only one reuse distance that is not $\infty$. So the average reuse distance in the preceding \textit{4}-length window is 2. The frequency of ``b'' in the past \textit{4} accesses is 2.

\subsection{Data Clustering Analysis}
In addition to the above locality intrinsic features, we design a feature through a novel data clustering analysis that clusters data accesses by locality patterns exhibited inside a cache trace. As previously computed, every data access has a data address delta. Therefore we are able to cluster data accesses based on data address deltas.
The clustering analysis groups distinct address deltas. Each group contains similar address deltas, corresponding to a group ID. As a result, the corresponding feature tensor of each data access contains a cluster ID by its address delta.

\textit{K-means}~\cite{MacQueen1967} is a classical algorithm to cluster data blocks based on a certain kind of data similarity. To facilitate the implementation of \textit{K-means} in the proposed data clustering analysis, we implemented an auto-partition algorithm that divides the entire data accesses into an appropriate number of clusters. As a result, the entire data accesses are as evenly distributed among those clusters.





\begin{figure}[t]
\centering
\includegraphics[width=0.85\linewidth]{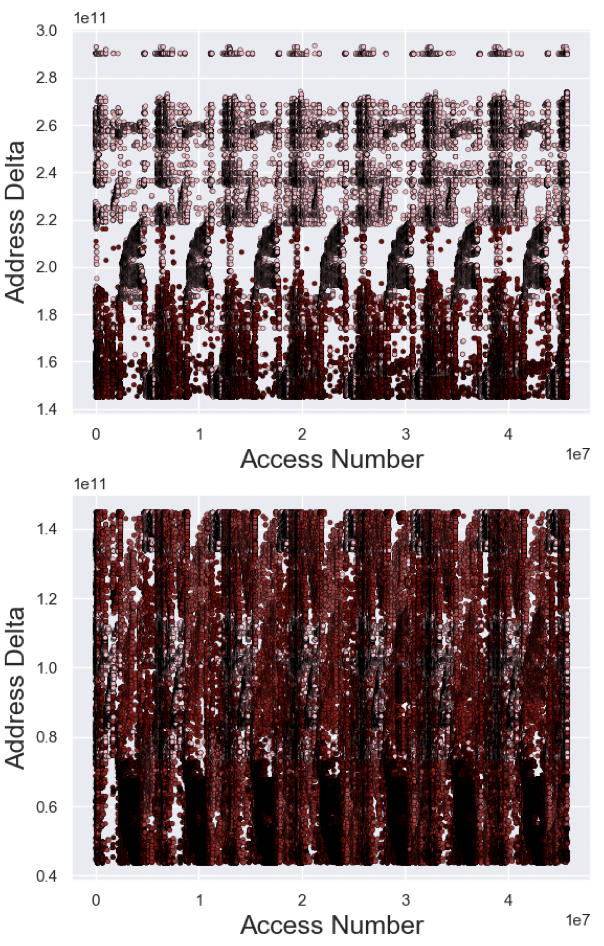}
\caption{Two of six clusters on the \textit{src1} trace. Data accesses are colored according to their raw data addresses.}
\label{fig:clustering}
\end{figure}

Figure~\ref{fig:clustering} shows two of six address-delta clusters of \textit{src1} trace. We color the points according to their raw address. Generally, address-delta distributions show pattern and phase behaviors. Different clusters have different address-delta distributions. Even in a single cluster, the same address-deltas may belong to different raw address. 


\subsection{From a Cache Trace to a Data Set}
\label{sec:sample-gen}
Other than existing works~\cite{Hashemi+:ICML18}, a data set to train the neural network in our work is generated based on only a cache trace, which we claim as a main novelty.
A data set is composed of a number of samples; each sample is a sequence of one-dimensional feature tensors of length \textit{sequence\_length}. Each feature tensor corresponds to a data access and contains all the features defined above. In a nutshell, the shape of a sample is a tensor of ($sequence\_length$, $6$), where $6$ features are presented. As a result, the shape of a data set is a tensor of $(\#samples$, $sequence\_length$, $6)$.

In Figure~\ref{fig:lstm-example}, each column composes a feature tensor.
For example, at access time $7$, a feature tensor is ($b, 2, 2, 2, 2, 2$). The sample at time $7$ is the composition of the last $4$ feature tensors, supposing \textit{sequence\_length} is $4$, i.e., (($a, 2, 1, 1.3, 3, 1$), $(b, 2, \infty, 2, 2, 1)$, $(a, 2, 2, 2, 2, 1)$, $(b, 2, 2, 2, 2, 2))$. The green box frames the sample at time $7$; the blue frames the next sample at time $8$. The generation of samples is a linearly iterative process in linear time complexity.





\section{Prediction-Based Pseudo OPT}
\texttt{OPT} discards the data block that will not be used for the longest time in the future. Forward reuse distance just tells us which data will be accessed the farthest.
After we have trained the neural network, we apply the network on-line through a cache trace, predicting the forward reuse distance of every access on-the-fly. With the predicted forward reuse distance for every access, we implement a prediction-based pseudo OPT policy. We call it \texttt{pOPT} for short. Note that we are only implementing a simple, pseudo OPT policy.

\begin{algorithm}[t!]
\small
\begin{algorithmic}[1]
\Require{TM} \Comment{A LSTM model that has been trained}
\Require{N} \Comment{Total number of accesses of a given trace}
\Require{Trace[1 \dots N]} \Comment{A given trace}
\Require{C} \Comment{Cache size}
\Require{Cache} \Comment{Cache is a hash map}
\Ensure{NumMiss} \Comment{Number of missed accesses}

\Function{pOPT}{}
\State $NumMiss \gets 0$
\For {$access\ i$ such that $0 < i \leq N$}
  \State $block \gets Trace[i]$
  \State $sample \gets \Call{GenerateSample}{i}$
  \State $val_{pred} \gets \Call{TM}{sample}$
  \State $fwd\_rd \gets \Call{ValueToForwardRD}{val_{pred}}$
  \If {$Cache[block] = 0$}
    \State $NumMiss \gets NumMiss + 1$
    \If {$len(Cache) = C$}
      \State \Call{EvictOneEntry}{}()
    \EndIf
  \EndIf
  \State $Cache[block] = i + fwd\_rd$ \Comment {Insert or update}
\EndFor
\State \Return $NumMiss$
\EndFunction

\Function{GenerateSample}{i}
\State /*Omit function body*/
\EndFunction

\Function{ValueToForwardRD}{v}
\State /*Omit function body*/
\EndFunction

\Function{EvictOneEntry}{}
\State $farthest\_time \gets 0$
\State $candidate \gets null$
\ForAll {$block$ in Cache}
  \State $next\_access\_time \gets Cache[block]$
  \If {$next\_access\_time > farthest\_time$}
    \State $farthest\_time \gets next\_access\_time$
    \State $candidate \gets block$
  \EndIf
\EndFor
\State {\bf del} $candidate$
\EndFunction

\end{algorithmic}
\caption{Prediction-based OPT}
\label{alg:popt}
\end{algorithm}

Algorithm~\ref{alg:popt} presents the \texttt{pOPT} policy. It is a pseudo OPT replacement policy. The input contains a cache trace of length $N$, a pre-trained model $TM$, and a cache of size of $C$. The output is the miss ratio for the given trace to run through the \texttt{pOPT} cache. We use a hash table to implement the cache. The key of the hash table is a cached data block. The value is the next access time for a cache entry, which is used for cache eviction. 

The \texttt{pOPT} function is the main entry. It iterates every data access in Line $3$. For each access, it computes the corresponding feature tensor on-the-fly by following the aforementioned computing procedures. Then it composes a sample and feeds it in the neural network to obtain a predicted forward reuse distance for the being accessed data block. The generation of a sample is discussed in Section~\ref{sec:sample-gen}. We omit its function body in the algorithm. Line $8$ examines if the being accessed data block is already in cache by examining if the next access time equals to 0. The next access time being 0 denotes the data block never being stored.
If the data block is not in cache, we increment the miss number. If the cache is full, Line $11$ evicts the cache entry that will not be accessed for the longest time in the future to insert current data block. Line $14$ inserts or updates the next access time of current block. The time complexity depends on the time complexity of an inference operation of the LSTM model.

\section{Evaluation}
This section evaluates the prediction accuracy of the proposed neural network mode and cache performance of the \texttt{pOPT} policy.

\subsection{Cache Policies}

\def \LRU {\texttt{LRU}}
\def \LFU {\texttt{LFU}}
\def \twoQ {\texttt{2Q}}
\def \LIRS {\texttt{LIRS}}
\def \ARC {\texttt{ARC}}
\def \OPT {\texttt{OPT}}
\def \POPT {\texttt{pOPT}}

\texttt{pOPT} is compared with 3 practical policies, \LRU, \twoQ~\cite{JohnsonS:VLDB94}, \ARC~\cite{MegiddoM:FAST03}, and 1 ideal policy, \OPT~\cite{Mattson+:IBM70}. \LRU\ captures \textit{data recency} by replacing the least recently accessed data blocks. It does not utilizes \textit{data frequency}. \LFU\ makes use of no data recency but data frequency, thus it may accumulate old data blocks in cache that are frequently accessed but no more used. Later, many cache policies strive to improve based on \LRU\ and \LFU. \texttt{LRU-K}~\cite{Oneil+:SIGMOD93} approximates \LFU\, while considering data recency by tracking the times of the last \textit{K} references to estimate reuse distances for references. Its implementation takes logarithmic time complexity. \twoQ\ mimics \texttt{LRU-2} but with constant time overhead; therefore, we compare \POPT \ with \twoQ.  
\ARC\ uses a learning rule to tune cache actions by balancing between data recency and frequency on-line and is known to be one of the best performing practical policies. \ARC\ is extensively used in production systems~\cite{Waldspurger+:USENIX17}. 
\OPT~\cite{Mattson+:IBM70} is an ideal policy that gives a theoretical upper-bound for cache optimization. \OPT\ takes precise future information to achieve optimal caching.

\begin{table*}[]
\scalebox{0.85}{
\begin{tabular}{|c|c|c|c|c||c|c|c|c|}
\hline
\textbf{Source} & \textbf{Domain (\# volumes)} & \textbf{Name} & \textbf{Trace Length} & \textbf{\#Data Blocks} & \textbf{Seq. Length} & \textbf{Training Size} & \textbf{Validation Size} & \textbf{Batch Size} \\ \hline
\multirow{13}{*}{\begin{tabular}[c]{@{}c@{}}MSR\\ (13 servers,\\ 36  volumes,\\ 179 disks)\end{tabular}} & Test web server (4) & \textit{wdev} & 3,024,140 & 162,629 & 512 & 1,000,000 & 10,000 & 8 \\ \cline{2-9}
 & Terminal server (1) & \textit{ts} & 4,181,323 & 256,922 & 512 & 1,000,000 & 10,000 & 8 \\ \cline{2-9} 
 & Research projects (3) & \textit{rsrch} & 3,508,103 & 279,128 & 1024 & 1,000,000 & 50,000 & 8 \\ \cline{2-9} 
 & Hardware monitoring (2) & \textit{hm} & 11,183,061 & 715,049 & 1024 & 5,000,000 & 10,000 & 32 \\ \cline{2-9} 
 & Firewall/web proxy (2) & \textit{prxy} & 351,361,438 & 842,095 & 256 & 10,000,000 & 50,000 & 64 \\ \cline{2-9} 
 & Source control (3) & \textit{src2} & 28,997,811 & 10,939,638 & 512 & 10,000,000 & 50,000 & 64 \\ \cline{2-9} 
 & Project directories (5) & \textit{proj} & 599,716,005 & 325,439,390 & 256 & 10,000,000 & 10,000 & 64 \\ \cline{2-9} 
 & Web/SQL Server (4) & \textit{web} & 78,662,064 & 20,563,955 & 64 & 10,000,000 & 100,000 & 64 \\ \cline{2-9} 
 & Web Staging (2) & \textit{stg} & 28,538,432 & 22,608,572 & 64 & 10,000,000 & 50,000 & 64 \\ \cline{2-9} 
 & Media Server (2) & \textit{mds} & 26,169,810 & 22,965,034 & 64 & 10,000,000 & 50,000 & 64 \\ \cline{2-9}
 & Print server (2) & \textit{prn} & 73,135,443 & 25,928,166 & 512 & 10,000,000 & 50,000 & 64 \\ \cline{2-9} 
 & Source control (3) & \textit{src1} & 818,619,317 & 63,864,930 & 1024 & 10,000,000 & 100,000 & 64 \\ \cline{2-9}
 & User home directories (3) & \textit{usr} & 637,227,335 & 231,421,475 & 4096 & 1,000,000 & 100,000 & 8 \\ \hline
\end{tabular}
 }
\caption{Trace characteristics and selected training parameter values. For training parameters, we only tuned for a few distinct values. Here is the best selected.}
\label{tbl:benchmark}
\end{table*}

\subsection{Implementation}
Our prototype has two parts: an offline training component and an on-line cache. The offline component is input with a cache trace. It has several modules. The locality feature module generates a feature trace where each data access is associated with the aforementioned five locality-oriented features. Then a clustering module uses the \textit{K-means}~\cite{MacQueen1967} algorithm to group all accesses based on their memory address. The two modules form a feature-vector trace that is used further to generate training samples, i.e., a data set. A sample concatenates \textit{sequence\_length} feature tensors and corresponds to an access. The offline training component trains our RNN model. We implemented a simulator for the on-line \POPT\ cache by following Algorithm~\ref{alg:popt}. The \POPT\ cache runs through a cache trace with predicting forward reuse distance on-line by the trained RNN.

We implemented simulators for \LRU, {\twoQ}~\cite{JohnsonS:VLDB94}, and {\ARC}~\cite{MegiddoM:FAST03}. The \twoQ\ implementation refers to the version used in lease cache work~\cite{Li+:ASPLOS19}.



\subsection{Workloads}
\label{sec:workload}
We tested upon a set of storage traces collected by Narayanan et al.~\cite{Narayanan+:TOS08}. These traces collect active disk block accesses and were profiled from 13 different production servers, such as source control, web staging, etc, in the Microsoft Cloud. Recent decent cache studies use these traces for evaluation~\cite{Wires+:OSDI14,Waldspurger+:FAST15}. Each server owns one or more volumes of disks. Table~\ref{tbl:benchmark} summarizes the characteristics on the 13 traces.

\vspace{0.75\baselineskip}
\noindent {\bf \textit{Training and validation sets.}}
We split a cache trace with a ratio of $8:2$. The first fraction is used for training and the second for validation. We may not use all the samples in each fraction because it takes a huge amount of time to train a large number of samples and the disk pressure is incredibly high when we set a long \textit{sequence\_length} (discussed later in Section~\ref{sec:mem-overhead}). For the training set, we use the last few samples; for the validation set, we use the first few samples.

\vspace{0.75\baselineskip}
\noindent {\bf \textit{Neural network training.}}
Table~\ref{tbl:benchmark} lists training parameters of our model, including training set size (number of training samples), \textit{sequence\_length}, validation set size and batch size. As tested, \textit{sequence\_length} is the most significant one and training set size is the second most significant. In addition to \textit{sequence\_length}, the proposed RNN model also has a handful of other parameters, such as \textit{batch\_size}, \textit{LSTM\_width}, \textit{LSTM\_layers}, \textit{epochs}, \textit{learning\_rate}, and \textit{dropout}. \textit{LSTM\_width} denotes the width of a LSTM cell~\cite{HochreiterS:1997}, \textit{LSTM\_layers} denotes how many LSTM layers we use, and \textit{dropout} denotes the drop ratio of weight of each neural network layer. For all workloads, we set total number of epochs as $1000$, \textit{learning\_rate} as 0.001, \textit{LSTM\_width} as 256, \textit{dropout} as 0.2, and \textit{LSTM\_layers} as 2. We will study the effect of these tunable parameters in Section~\ref{sec:hyper-effect}. 

All deep learning tasks were performed on a Nvidia Tesla V100 GPU~\cite{nvidia:v100} card of the recent Volta architecture, which has 5120 streaming cores, 640 tensor cores and 32GB memory capacity. The CPU host is Intel Xeon CPU 8163 2.50GHz, running Linux kernel 5.0. All the other non-deep learning tasks were run on the host.



\subsection{Prediction Accuracy}

\begin{figure*}[t!]
\centering
\subfloat{
  \includegraphics[width=\textwidth]{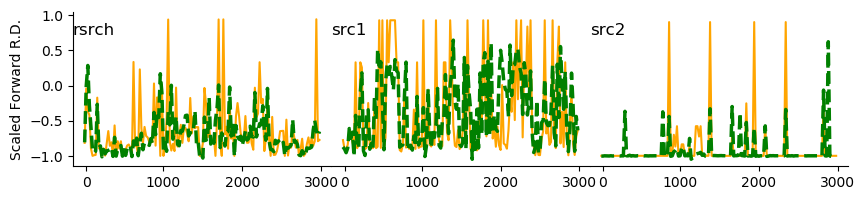}
  \label{fig:rsrsh-1-pred}
}
\vspace{0em}
\\
\subfloat{
  \includegraphics[width=\textwidth]{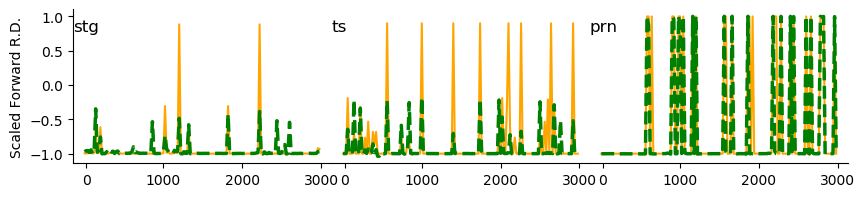}
  \label{fig:rsrsh-1-pred}
}
\vspace{0em}
\\
\subfloat{
  \includegraphics[width=\textwidth]{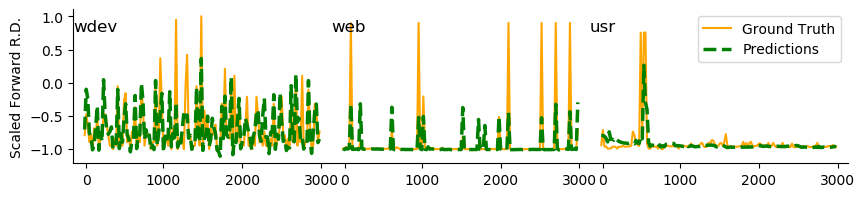}
  \label{fig:rsrsh-1-pred}
}
\caption{Prediction results of our RNN model. X-axis is the access time. We show the first 3000 data accesses in the validation set. Y-axis is the scaled forward reuse distance. In this figure, $-1$ denotes $\infty$ reuse distance.}
\label{fig:all-preds}
\end{figure*}



We pre-compute the forward reuse distance for all accesses off-line as the ground truth. Figure~\ref{fig:all-preds} depicts the prediction accuracy results, by comparing predictions to the ground truth, of 9 traces with distinct phases behaviors. From a good visual perspective, we show the comparison results of 3000 data accesses in the validation set. The overall prediction trend matches very well with that of the ground truth, especially in \textit{rsrch}, \textit{src1} and \textit{wdev}, though the two absolute values have slight difference for some accesses.

The ground-truth values vary rapidly and aggressively from time to time in \textit{rsrch}, \textit{src1} and \textit{wdev}. Our predictor demonstrates a high degree of prediction matching upon the variance trend. In particular, we have the best prediction accuracy visually in \textit{wdev}. Other traces, \textit{src2}, \textit{stg}, \textit{ts}, and \textit{web} show many target spikes, meaning that a non-zero reuse distance appears once in a while. Our predictor is able to catch most of the spikes, though it causes mis-predictions of spikes, for example in \textit{src2} and \textit{stg}. The last trace \textit{usr} is special, with a single spike and many small yet rapid varying target values. It is worth noting that \textit{hm}, \textit{prn}, \textit{mds}, \textit{proj} and \textit{prxy} have a similar reuse-distance pattern. They only have reuse distance of $1$ and $\infty$, that is, a data access is either reused in the next access or never reused. We show only the results of \textit{prn} to represent all of them.

There are two potential reasons for that the predictor has false positives and the absolute prediction value is not 100\% the same as the ground truth. First, the training set is not large enough, so the model may not capture all behaviors. Second, the set of training parameters was not best selected. However, we would argue that there is an exponential number of combinations of values for the training parameters. It is unrealistic to manually enumerate and test all of them. In the future, we would seek to use the advanced learn-to-learn~\cite{Golovin+:KDD17} techniques to auto-tune the best set of training parameters. Moreover, even if we do not have completely precise prediction, our predicted results of forward reuse distance are quite useful, which will be seen next in the evaluation of \POPT\ policy.


\begin{figure*}[bhpt]
\centering
\subfloat[MRCs on \textit{hm}.]{
  \includegraphics[width=0.32\textwidth]{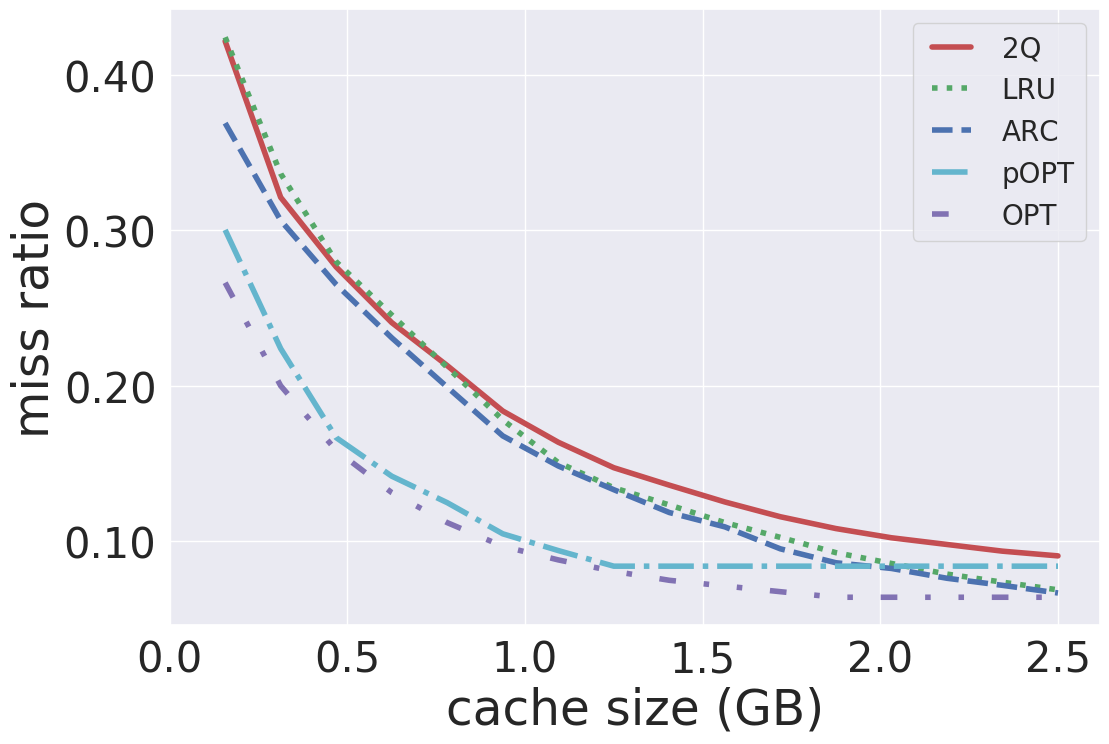}
  \label{fig:rsrsh-1-pred}
}
\subfloat[MRCs on \textit{mds}.]{
  \includegraphics[width=0.33\textwidth]{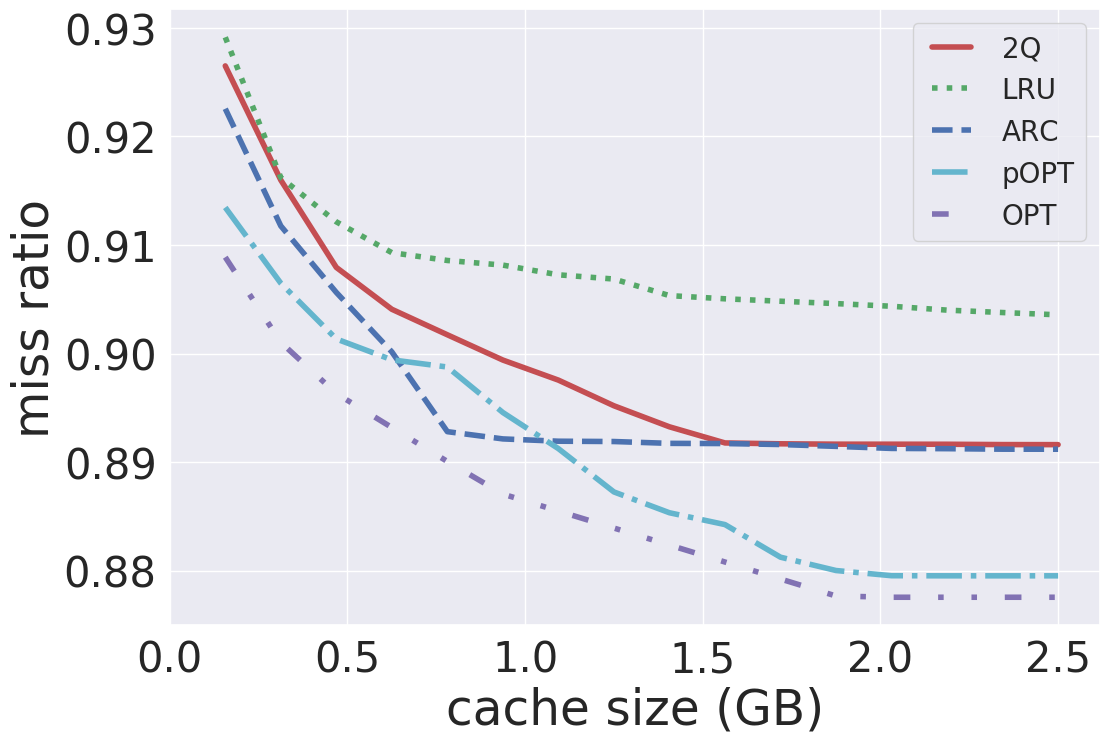}
  \label{fig:rsrsh-1-pred}
}
\subfloat[MRCs on \textit{prn}.]{
  \includegraphics[width=0.32\textwidth]{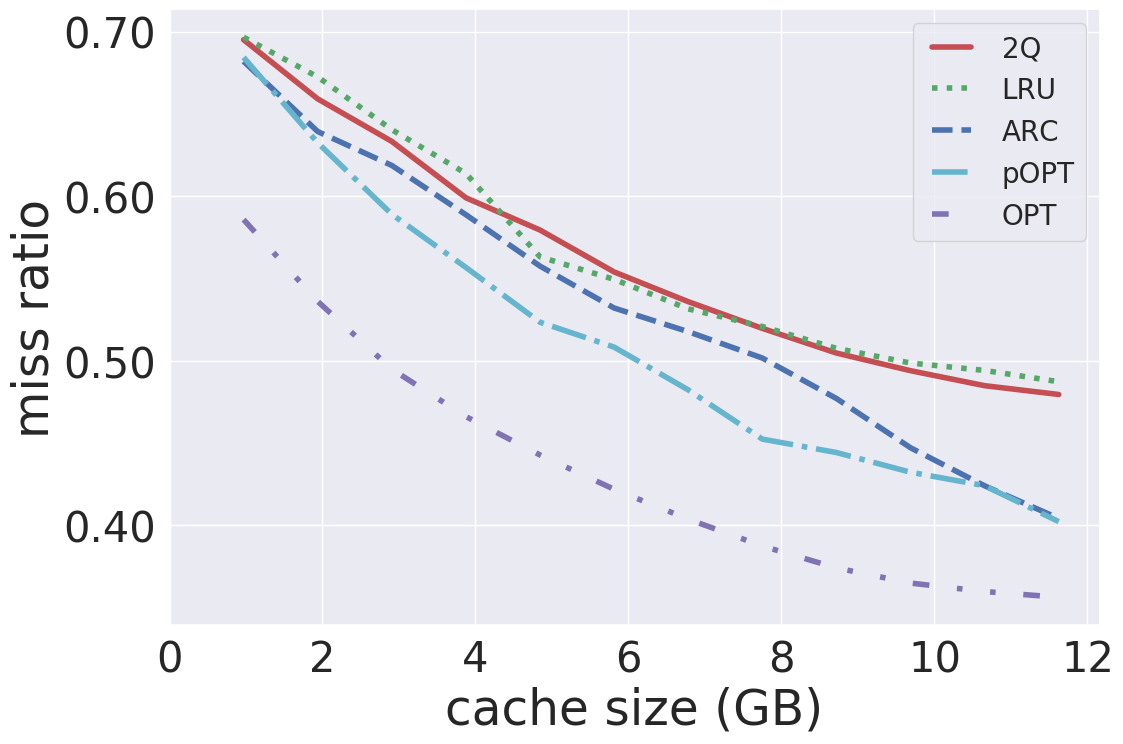}
  \label{fig:rsrsh-1-pred}
}
\\
\subfloat[MRCs on \textit{proj}.]{
  \includegraphics[width=0.32\textwidth]{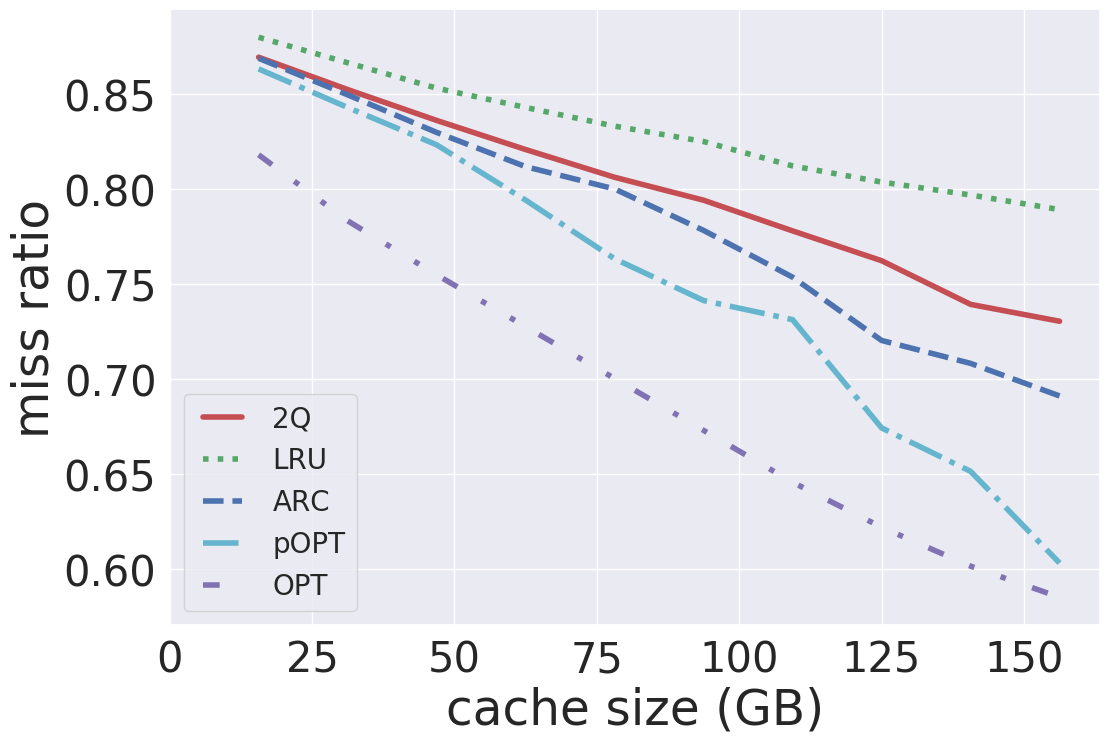}
  \label{fig:rsrsh-1-pred}
}
\subfloat[MRCs on \textit{prxy}.]{
  \includegraphics[width=0.32\textwidth]{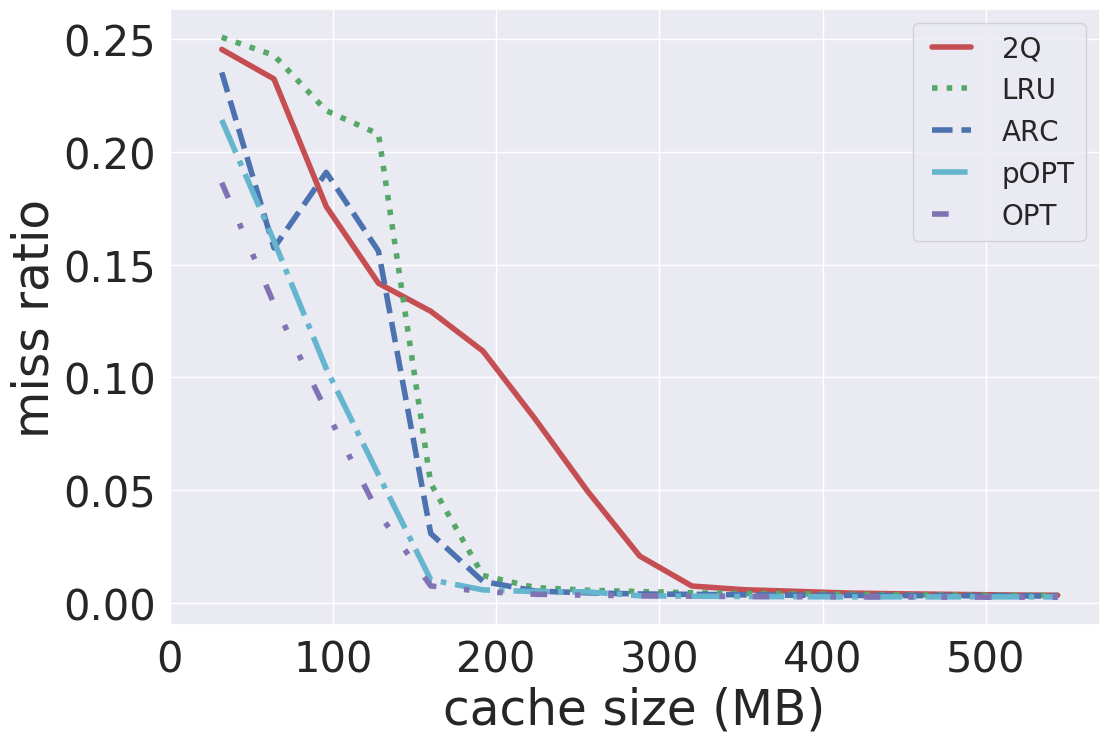}
  \label{fig:rsrsh-1-pred}
}
\subfloat[MRCs on \textit{rsrch}.]{
  \includegraphics[width=0.32\textwidth]{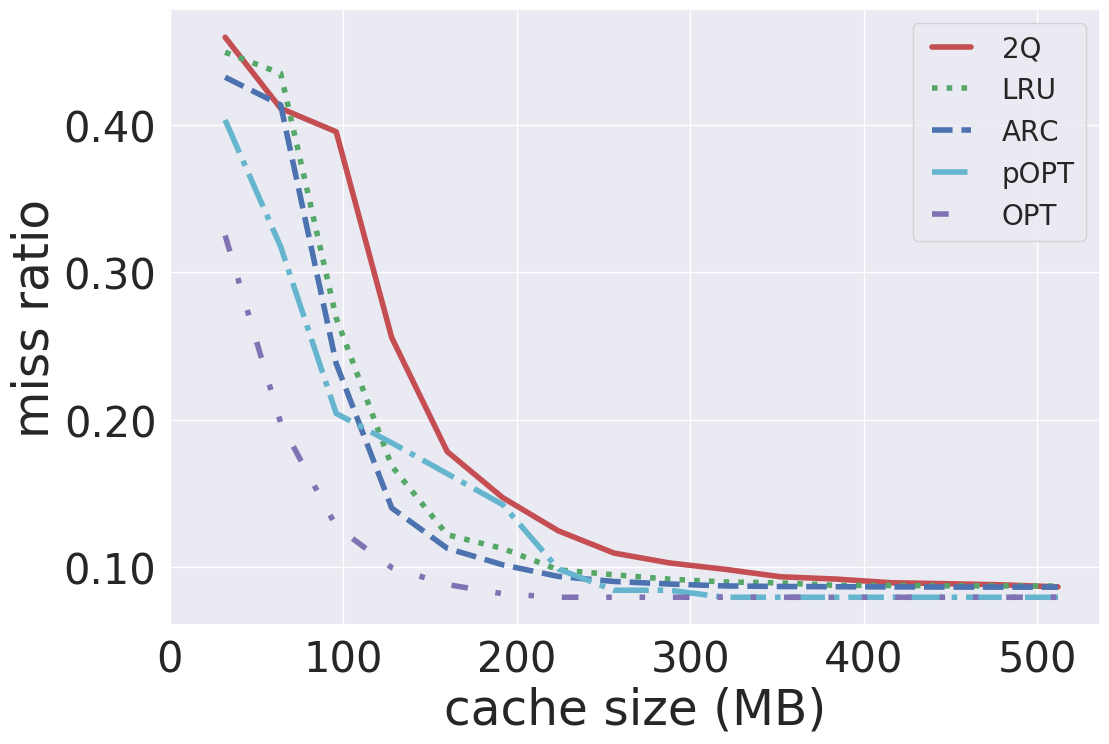}
  \label{fig:rsrsh-1-pred}
}
\\
\subfloat[MRCs on \textit{src1}.]{
  \includegraphics[width=0.32\textwidth]{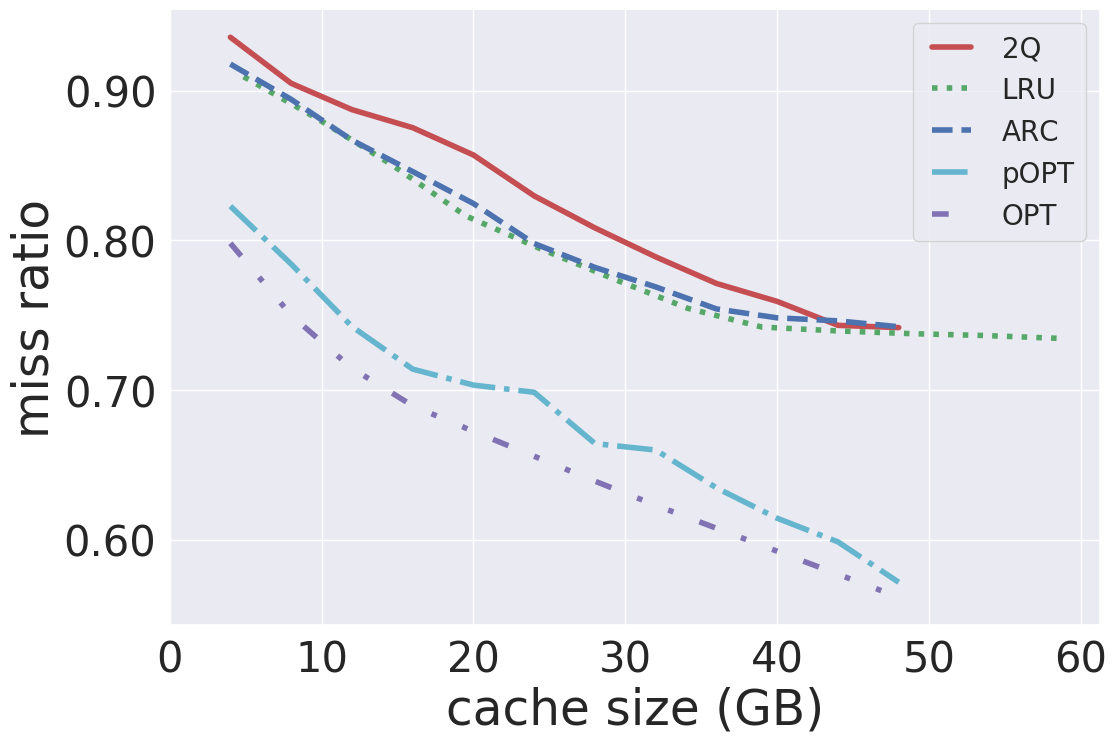}
  \label{fig:rsrsh-1-pred}
}
\subfloat[MRCs on \textit{src2}.]{
  \includegraphics[width=0.32\textwidth]{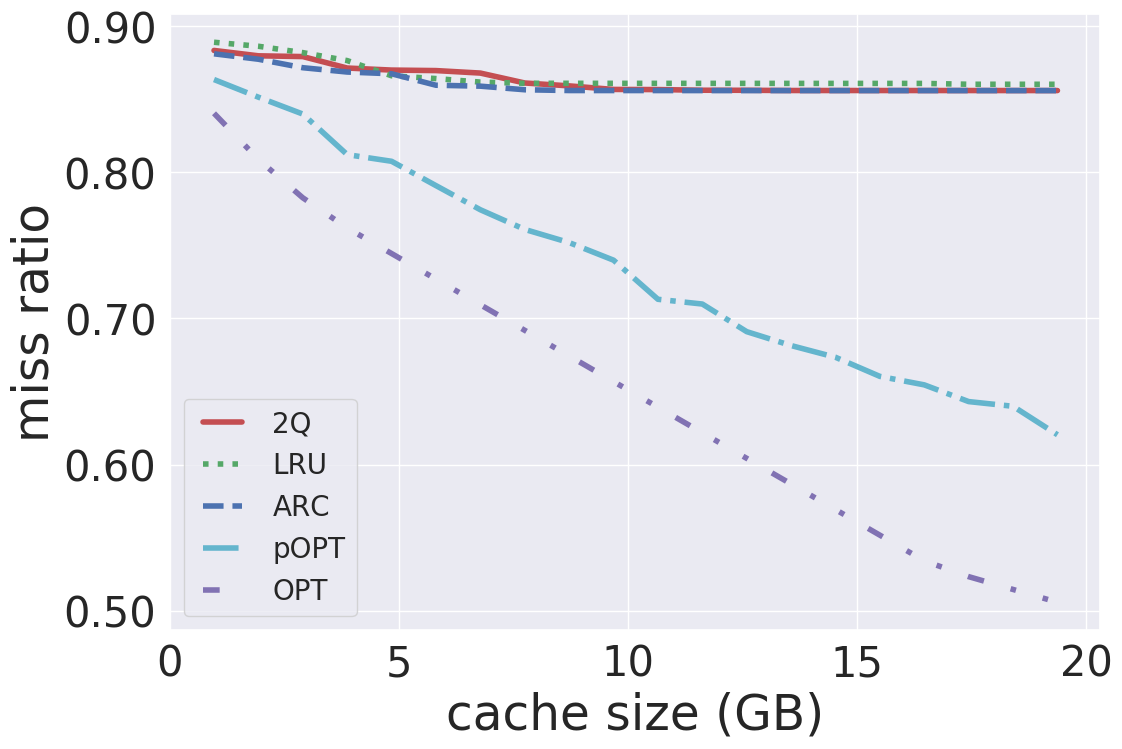}
  \label{fig:rsrsh-1-pred}
}
\subfloat[MRCs on \textit{stg}.]{
  \includegraphics[width=0.32\textwidth]{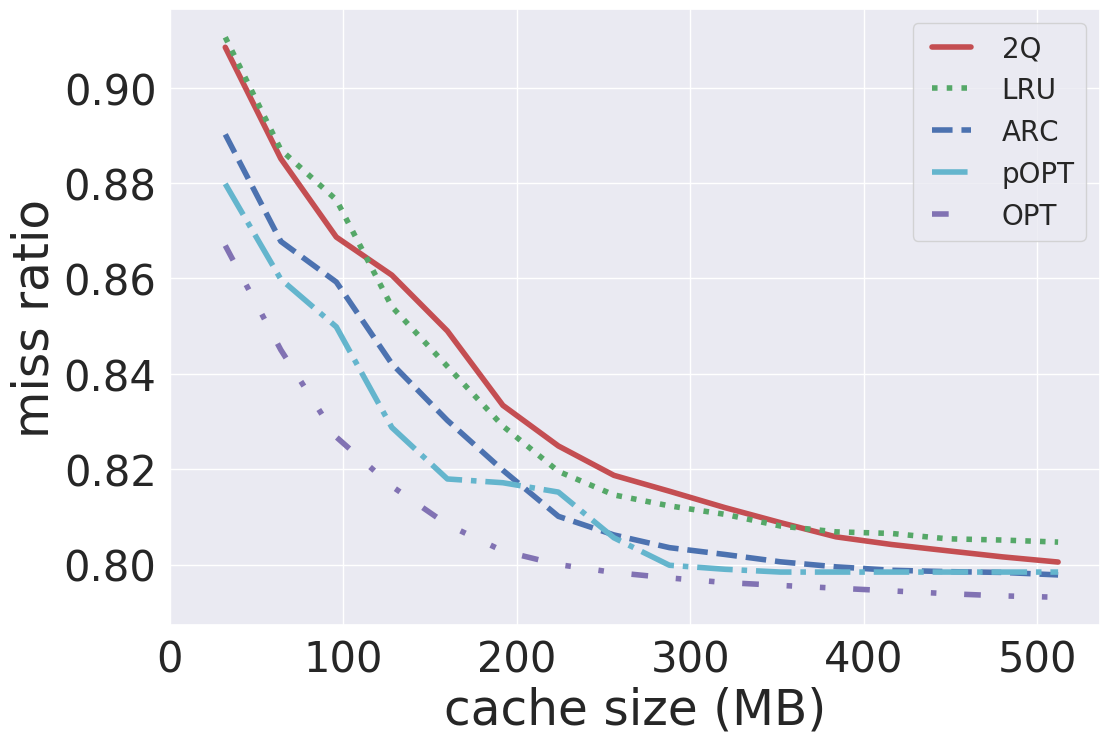}
  \label{fig:rsrsh-1-pred}
}
\\
\subfloat[MRCs on \textit{ts}.]{
  \includegraphics[width=0.32\textwidth]{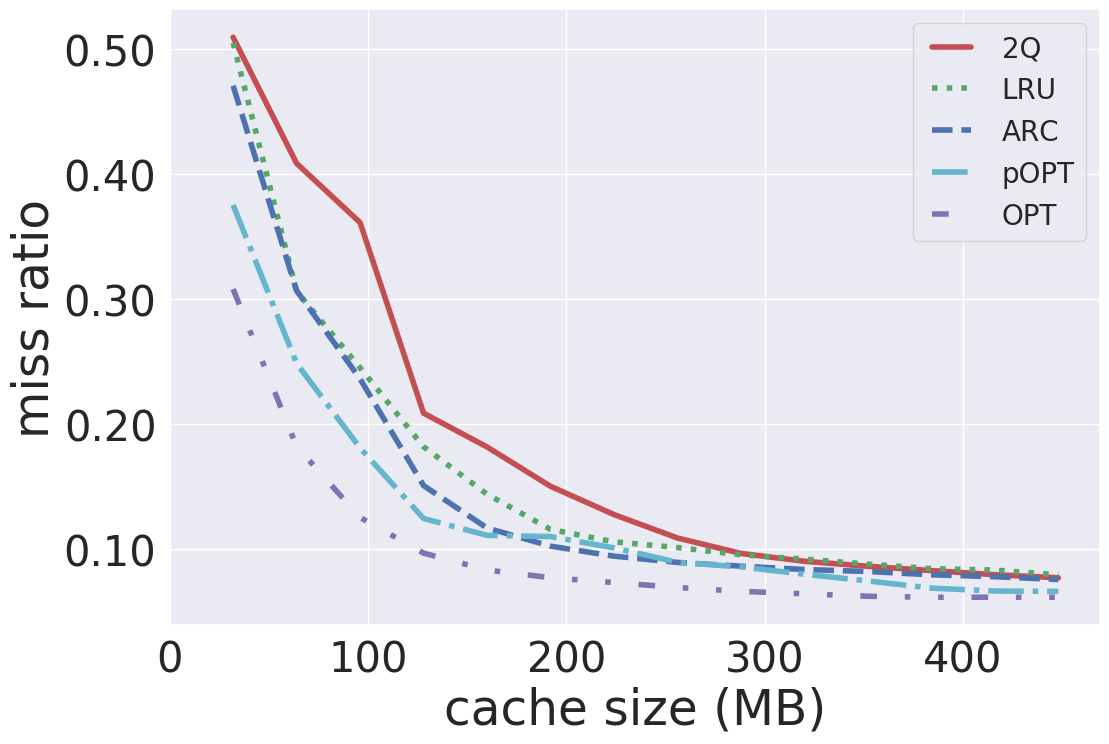}
  \label{fig:rsrsh-1-pred}
}
\subfloat[MRCs on \textit{wdev}.]{
  \includegraphics[width=0.32\textwidth]{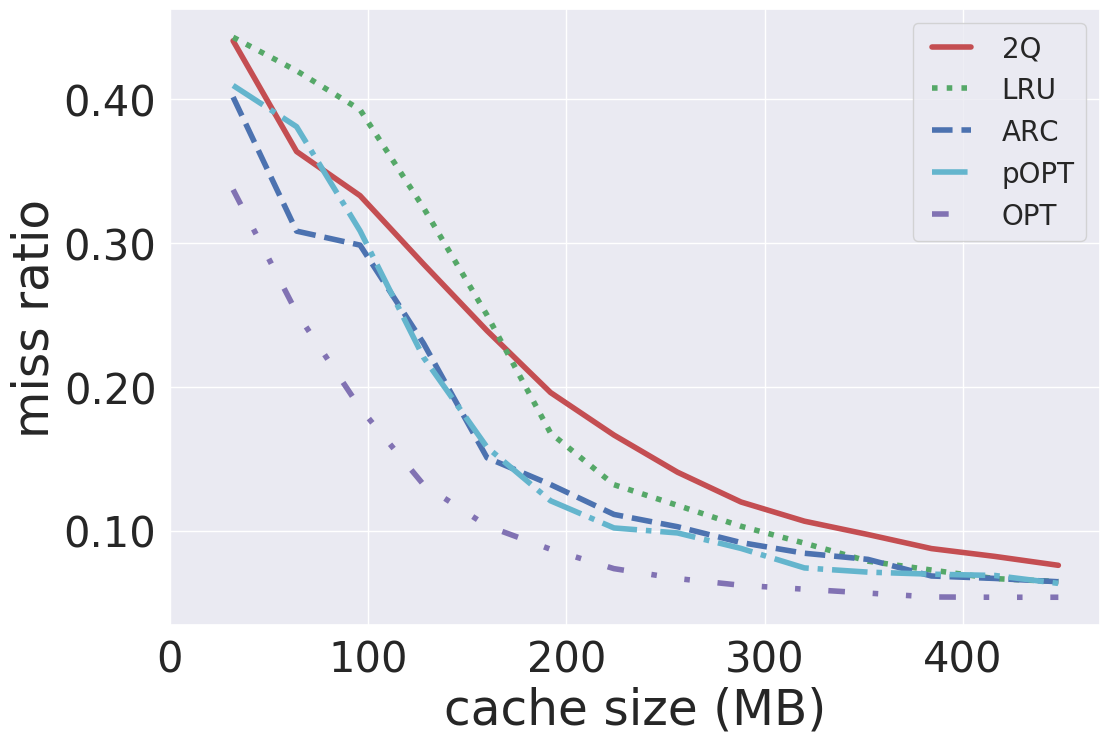}
  \label{fig:usr-0-64}
}
\subfloat[MRCs on \textit{web}.]{
  \includegraphics[width=0.32\textwidth]{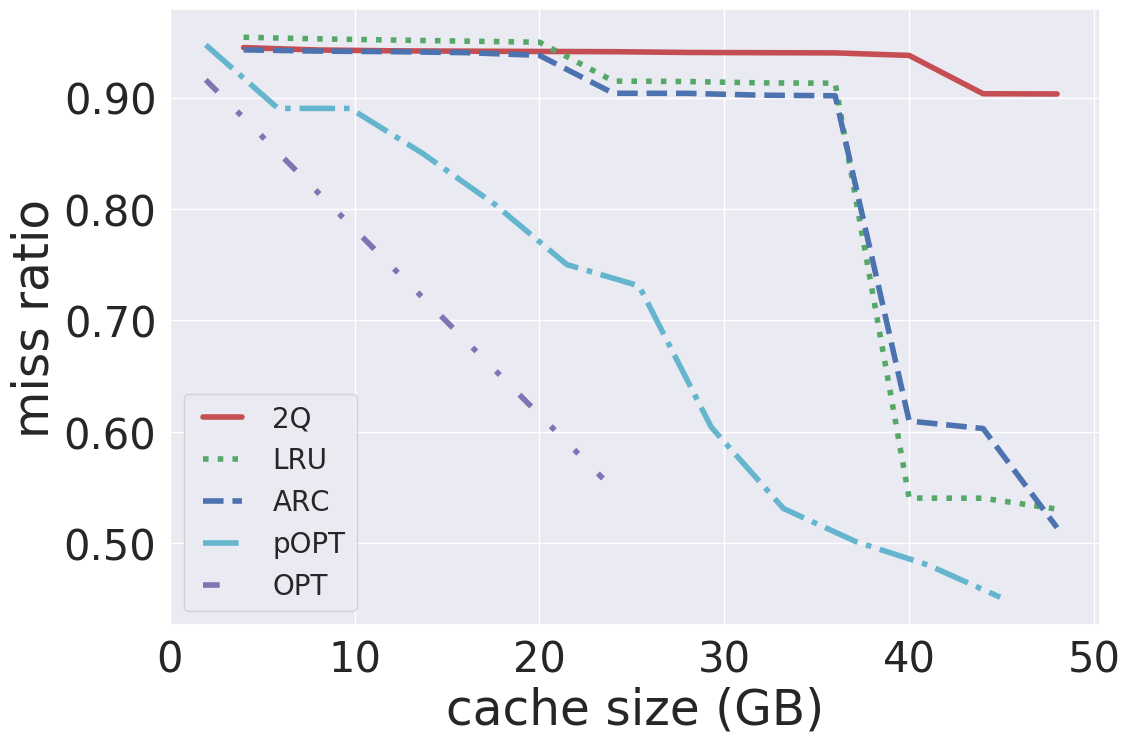}
  \label{fig:usr-0-1024}
}
\\
\subfloat[MRCs on \textit{usr}.]{
  \includegraphics[width=0.32\textwidth]{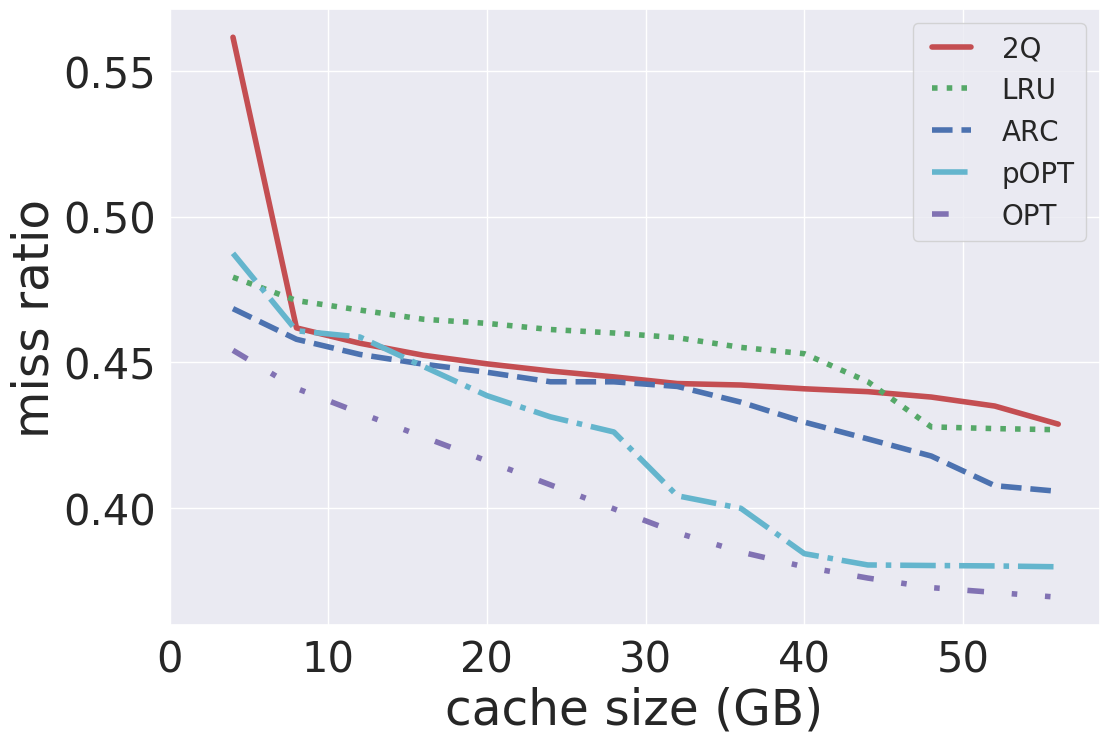}
  \label{fig:rsrsh-1-pred}
}
\subfloat{
  \includegraphics[width=0.63\textwidth]{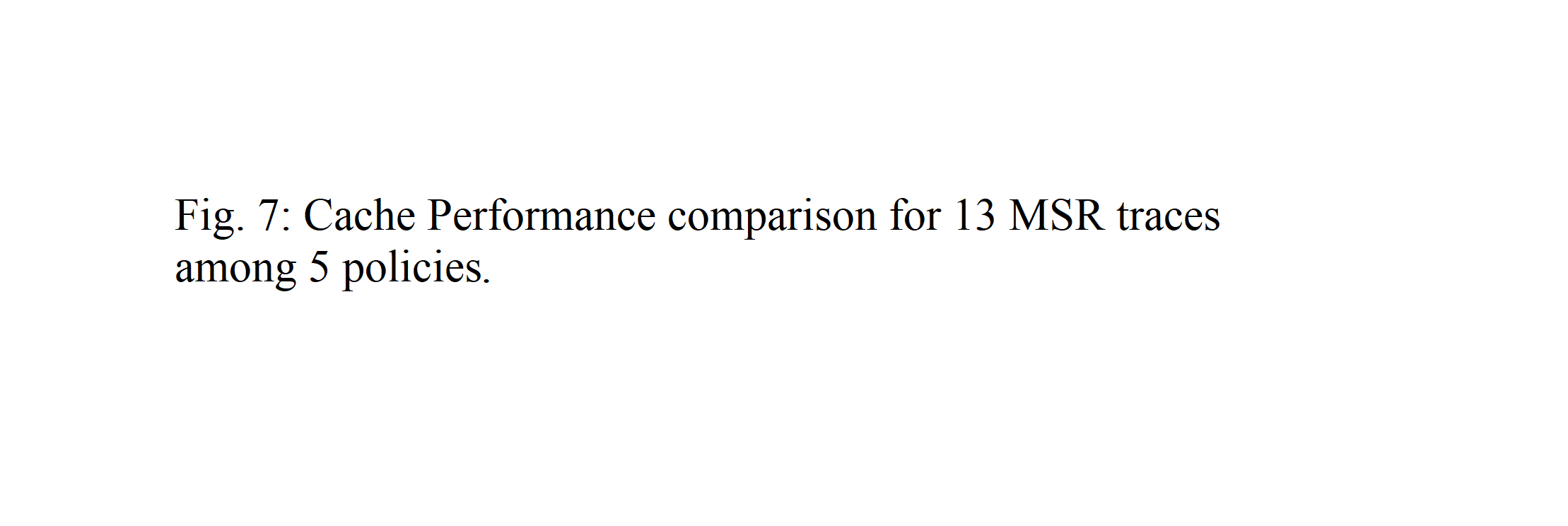}
}
\label{fig:all-cache}
\end{figure*}

\subsection{Performance of \POPT\ Cache}
Figure 7 compares performance for 5 cache replacement policies. Each graph shows the miss ratio curves (MRC) of 5 policies. \OPT, the optimal policy, uses precise future knowledge, whereas \POPT\ uses predicted future knowledge. It is interesting that \POPT\ almost performs similar as \OPT\ in 3 programs, \textit{hm}, \textit{prxy} and \textit{src1}. In \textit{prn}, \textit{proj} and \textit{src2}, there is a big gap between \POPT\ and \OPT. In \textit{mds}, \textit{rsrch}, \textit{ts} and \textit{stg}, \POPT\ is consistently a little worse than \OPT. In all tests, \POPT\ has 2.3\% higher miss ratio than \OPT\ on average.
In \textit{hm}, \POPT\ starts with a close miss ratio to \OPT\ until $1.2$GB cache size, and then stays the same at the lowest point, showing that \POPT\ performs poor in achieving the best miss ratio when cache size is extremely large.
In several traces, \textit{mds}, \textit{proj}, \textit{rsrch}, \textit{src1} and \textit{stg}, the gradient of the MRC gets smaller and larger in an interleaving manner when cache size increases. In contrast, the MRCs on \OPT\ are always concave curves, suggesting that predictions of forward reuse distance certainly do not tell the precise future and hence brings unstable performance gain when cache increases. 

\POPT\ is better than \LRU, \twoQ\ and \ARC\ by 13.7\%, 19.2\% and 8.6\% on average across all tests.
\POPT\ clearly outperforms three practical policies, \LRU, \twoQ\ and \ARC\ in traces \textit{proj}, \textit{prxy}, \textit{src1}, \textit{src2} and \textit{ts}. \POPT\ is always better than \twoQ\ regardless of cache size. \POPT\ is better than \LRU\ on all traces for all cache sizes but \textit{rsrch}. In \textit{rsrch}, \POPT\ starts with a low miss ratio than \LRU, then becomes a little worse when cache size ranges from 200MB to 240MB, and finally becomes better. \textit{rsrch} is special in that it has many short reuse distances, which is good for \LRU\ to capture data reuse. However, the prediction based on \POPT, for some cache size, may unexpectedly thrash data in cache. To further understand the reason, it is the easiest to consider an access to a data block that has a long forward reuse distance is predicted with a short distance. When cache size is too small, the data block is evicted because the short prediction still can not keep the data block in cache; when cache size is too large, the data block is kept in cache regardless of the length of its forward reuse distance. In fact, in \textit{rsrch}, \LRU\ is close to \OPT\ comparing to in other traces. \POPT\ is better than \ARC\ on all traces except \textit{mds}, \textit{rsrch} and \textit{stg}. The average speedup of \POPT\ over \ARC\ is 8.6\%.

Among all tests, \textit{prxy} has the highest data reuse and \textit{stg} has the lowest, on average 437 and 1.1 accesses per data block, respectively. Results show that \POPT\ performs better on both cases. In \textit{prxy}, it outperforms \twoQ, \LRU\ and \ARC\ by 49.3\%, 26.0\% and 5.8\%, and is worse than \OPT\ by 14.0\% on average. In \textit{stg}, it outperforms \twoQ, \LRU\ and \ARC\ by 1.7\%, 1.7\% and 0.5\%, and is worse than \OPT\ by 1.1\%. It is instructive to see how a cache policy handles single-use data blocks. \OPT\ is aware of which data block is single-use. Our method predicts single-use data. In real-world applications, the prediction of single-use data blocks is actually very useful. For example, in a Cloud cache, if aware of which data block is single-use, we would prevent a cache from being polluted by it, which makes us save a considerable amount of machine resource and eventually money.

\subsection{The Effect of Hyper-Parameters}
\label{sec:hyper-effect}

\setcounter{figure}{7}    
\begin{figure}[t]
\centering
\subfloat[\textit{sequence\_length} is 64.]{
  \includegraphics[width=0.5\textwidth]{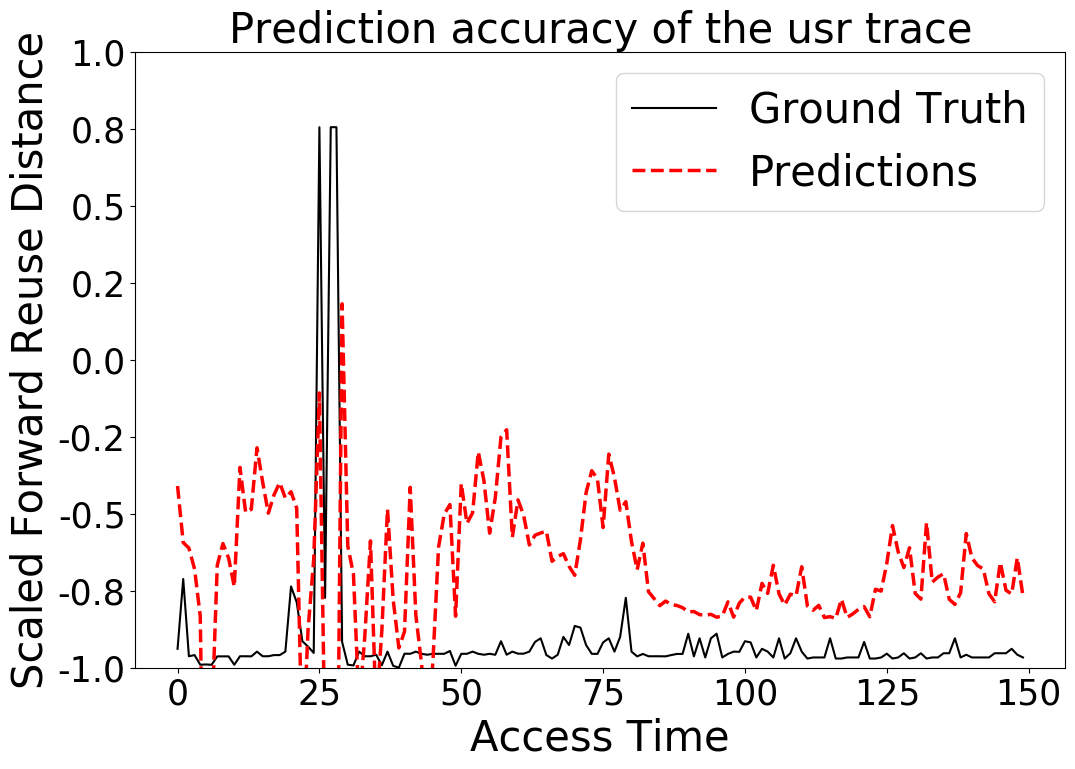}
  \label{fig:usr-0-64}
}
\subfloat[\textit{sequence\_length} is 1024.]{
  \includegraphics[width=0.5\textwidth]{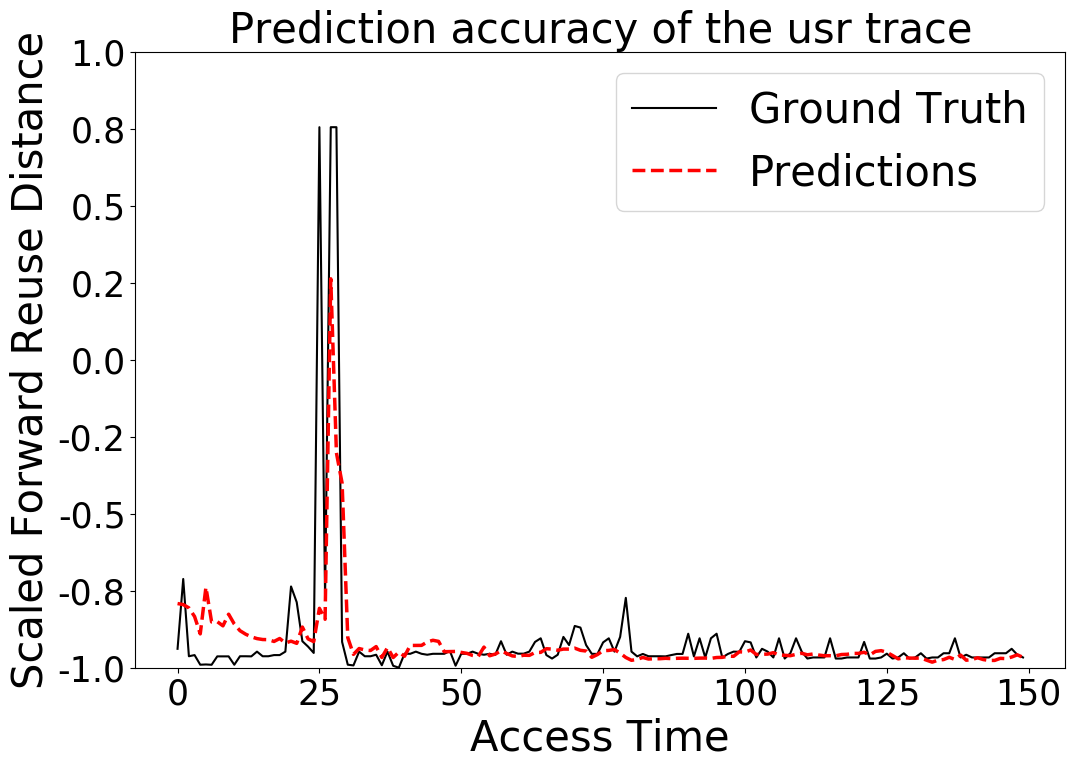}
  \label{fig:usr-0-1024}
}
\caption{Compare the prediction accuracy on \textit{usr} using different \textit{sequence\_length}s.}
\label{fig:seq-effect}
\end{figure}

The \textit{sequence\_length} controls how much history information LSTM looks back when predicting future. Larger \textit{sequence\_length} lets LSTM memorize more past program behaviors and be more precise in prediction. Figure~\ref{fig:seq-effect} clearly shows that larger \textit{sequence\_length} improves prediction accuracy from a comparison of the results of two different \textit{sequence\_length}s on \textit{usr}. The other training parameters stay the same in the two cases.




\begin{figure*}[t!]
\centering
\subfloat[Different \textit{batch\_size}s on \textit{wdev}.]{
  \includegraphics[width=0.49\textwidth]{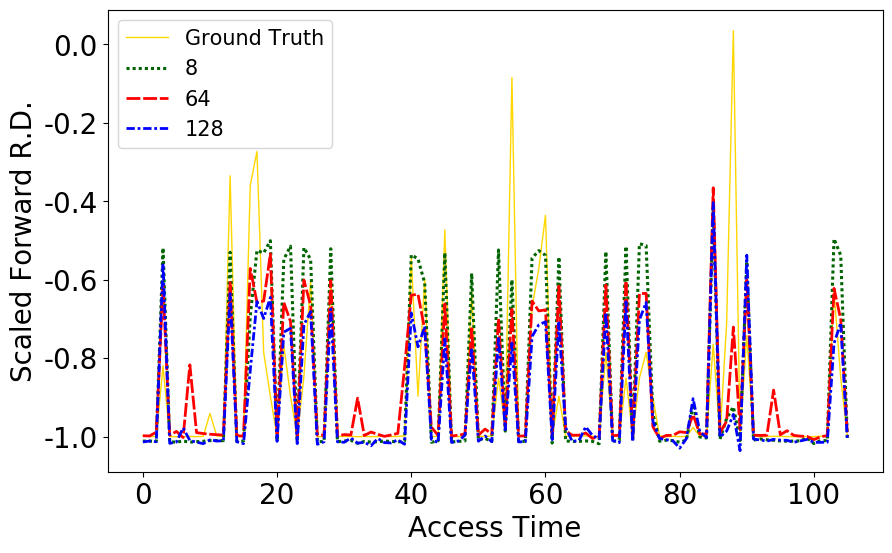}
  \label{fig:rsrsh-1-pred}
}
\subfloat[Different \textit{learning\_rate}s on \textit{wdev}.]{
  \includegraphics[width=0.49\textwidth]{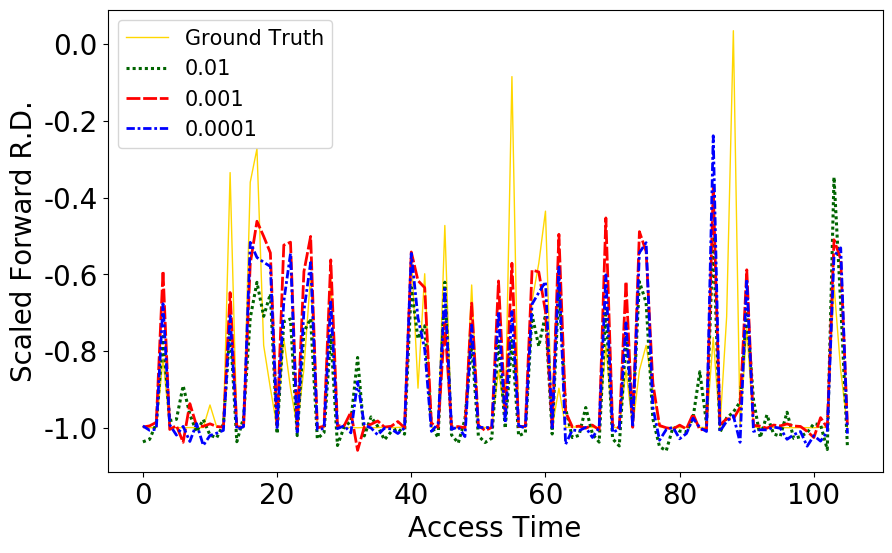}
  \label{fig:rsrsh-1-pred}
}
\caption{Compare the prediction accuracy on \textit{wdev} using different \textit{learning\_rate}s and \emph{batch\_size}s. We sample 100 points for the visual purpose.}
\label{fig:other-effect}
\end{figure*}

As mentioned in Section~\ref{sec:workload}, in addition to \textit{sequence\_length}, our model has \textit{LSTM\_width}, \textit{LSTM\_layers}, \textit{batch\_size}, \textit{learning\_rate}, \textit{dropout}, and \textit{epochs}. 
Tuning these parameters indeed improves prediction accuracy. We manually pick three or four discrete values for each parameter to study its effect. Figure~\ref{fig:other-effect} shows the prediction results of three different \textit{batch\_size}s and \textit{learning\_rate}s. When \textit{batch\_size} becomes small, the model is capable to predict large values, because small \textit{batch\_size} keeps the model updated quickly and converged well. For \textit{learning\_rate}, 0.01 is too large to find a good local optimal. While, 0.001 and 0.0001 perform similarly in terms of prediction accuracy.

We also found that larger values of \textit{LSTM\_layers} than 2 and larger values of \textit{LSTM\_width} than 256 resulted in no clear discrepancy with respect to prediction. 
Hence, we set \textit{LSTM\_layers} with 2 and \textit{LSTM\_width} with 256 in our study. Smaller \textit{dropout} performs better based on a comparison of dropout of 0.2, 0.5 and 0.7. For epochs, we set a very large number before training and stop training when validation results stay not improved for a few epochs. For limit of space, we omit the prediction figures for tuning these parameters in this paper.

\subsection{Training Time and Memory Overhead}
\label{sec:mem-overhead}
A cache trace often contains millions or billions of accesses. A few samples are not sufficient to train a model well. Long reuse distance happens in a cache trace, as shown by Figure~\ref{fig:trace-pattern}. Short \textit{sequence\_length} is not good enough to capture entire data reuses. Therefore, to solve our prediction problem, a large data set is needed. 

The data set size is proportional to the size of training set $N$, the size of validation set $M$ and \textit{sequence\_length}. A single feature vector has around 48 bytes. The total data set size is $48 \times \textit{sequence\_length} \times (N+M)$. Take \textit{hm} for example, $N=5,000,000$, $M=10,000$, and $\textit{sequence\_length} = 1024$. The data volume is $251$GB. This huge data set takes 3.45 days to run, as tested, just for one set of parameters. Therefore, we are not able to find the best set of parameters in reasonable time. More specifically, consider that for each parameter we randomly pick 3 values. For 13 workloads, we have in total 9477 trials. As tested, each trial takes almost one day to run, the total time is near for ever.

These numbers indirectly suggest the hardness of the problem studied in this paper. Fortunately, our results have shown the effectiveness of our model in limited exploration of parameter search space. Storage pressure and long training time are almost unavoidable if we would like to deal with more samples and let LSTM look back at more history information, i.e., larger \textit{sequence\_length}. Therefore, training our model not only requires to select the best set of parameters, but also needs to compromise a data set explosion problem, thus making the prediction of forward reuse distance very challenging.

\subsection{More Discussions}
The prediction of forward reuse distance  is a critical caching problem, which cannot be solved with conventional caching algorithms easily. RNN has demonstrated its success in machine translation problems. Bridging the gap from RNN to the prediction of forward reuse distance makes one of the contributions of this paper. This paper is the first work that manages to verify that applying LSTM onto the problem of forward reuse distance prediction is viable. 

In addition to the verification, a lesson with respect to practicality is learnt from this study. Training a dense RNN is time costly and takes days even with a moderate data set when we are testing upon a single advanced Nvidia V100 GPU card. Therefore, for practical deployment of RNNs for solving systems problems, a distributed training approach needs to be explored~\cite{Verbraeken+arxiv19}. Moreover, search space of optimizing hyper-parameters is quite huge. To figure out the best set of parameters in reasonable time, AutoML~\cite{Golovin+:KDD17} should be leveraged. Distributed training or AutoML for practical deployment is not the focus of this paper. 
Instead, this paper, as a theoretical study, has managed to verify the effectiveness of the application of LSTM on a hard caching problem.



\section{Related Work}

Jim{\'e}nez and Lin~\cite{JimenezL:HPCA01} designed a perceptron based branch predictor, which uses a linear classifier to predict if a branch is taken or not. It is one of the earliest works that apply machine learning techniques in architecture or systems problems. The perceptron is almost the simplest possible neural networks, which enables hardware implementation. Because of its simplicity, the perceptron technique is used later for several other systems problems~\cite{Teran+:MICRO16,LeventhalF:ICCD06}.
LSTM is a recurrent neural networks that is getting popular in the application in systems problems. Zekany et al.~\cite{Zekany+:MICRO16} applied an LSTM model in the compiler to find the hot path. The features are extracted from program semantics. The training of the neural network is run in a static compilation pass to predict run-time behaviors. 

Hashemi~\cite{Hashemi+:ICML18} also employed an LSTM model to learn object access patterns for optimizing the micro-architecture prefetching efficiency. Their goal is different than ours. Their work predicts the address of the next data access. Our work predicts when the next re-access of current accessing data block happens.  Similarly, their approach is also based on a memory access trace. Because they target hardware cache, program counters and memory address are both used as a feature for training. However, later work~\cite{Shi+:MICRO19} found that program counter is the most significant feature. With only using program counter information in the model, they achieve similar prediction accuracy.
In addition to the perceptron and LSTM models, Ipek~\cite{Ipek+:ISCA08} uses a reinforcement learning model to optimize a memory controller scheduling performance. Peled~\cite{Peled+:ISCA15} uses bandits to approximate semantic locality for a software prefetch optimization. 

Cache replacement policy is a popular research topic for the applications of deep learning models, because a time-series access trace can be easily fit into a set of neural networks, for example RNN.
Glider~\cite{Shi+:MICRO19} is the most recent research with respect to this topic. They used an attention-based LSTM neural network to discover the program insight and then designed hardware implementation using a simple vector machine. They feed in a memory access to the neural network. For their tests, they found that the model accuracy is actually highly related with program counter rather than memory address. Therefore, they use a cache trace that only consists of program counters. Their work explains the essential reason why the work~\cite{Hashemi+:ICML18} is effective, because it also used program counter as a feature.
DeepCache~\cite{Narayanan+:NETAI18} used LSTM to predict object popularity for Web content caches. Different than our approach, it also uses a complicated encoder-decoder RNN. Teran~\cite{Teran+:MICRO16} applied the perceptron learning to predict data reuse for optimizing the last-level cache due to its simplicity.


The optimal policy is MIN given by Belady~\cite{Belady66}. Mattson et al. developed the OPT stack algorithm which simulates Belady's optimal replacement for all cache sizes in two passes~\cite{Mattson+:IBM70}. The high cost of OPT stack simulation issue was addressed by Sugumar and Abraham, which used look-ahead and stack repair algorithm to avoid two-pass processing; moreover, grouping and tree lookup (instead of linear lookup) bring considerable speedup for stack simulation~\cite{SugumarA:SIGMETRICS93}.  The asymptotic cost per step is logarithmic in the number of groups. A recent theoretical study~\cite{Li+:ASPLOS19} demonstrated superior cache performance beyond OPT based on a series of higher-order cache or memory theories~\cite{Li+:ASPLOS19,Luo+:PPOPP17,Li+:ISMM14,Li+:ISMM16,Li+:IPDPS17,DingL:MSPC14,LiD:MSPC13,Luo+:ISPASS16,Luo+:PPOPP16}.

For hardware caches, Jain and Lin developed a policy called Hawkeye~\cite{JainL:ISCA16}.  Hawkeye keeps a limited history (a time window of 8x the cache size), uses interval counting (to target a single cache size), and leverages associativity and set dueling~\cite{Qureshi+:ISCA07} to compute OPT efficiently with low time and space cost in hardware.  In comparison, scaled-down simulation uses spatial sampling in software~\cite{Waldspurger+:USENIX17}.

\section{Conclusion}
In this paper, we have introduced the first use of deep learning to predict forward reuse distance. We have presented an LSTM-based RNN and a locality driven feature design to achieve high prediction accuracy. Based on the prediction results from the neural network, we have implemented a pseudo OPT-like replacement policy. In an offline testing, the new policy is demonstrated to outperform three practical state-of-the-art policies by up to 19.2\% and achieve close performance to OPT.



\ifCLASSOPTIONcaptionsoff
  \newpage
\fi

\end{document}